\documentclass{article} %
\usepackage{iclr2027_conference,times}

\usepackage{amsmath,amsfonts,bm}

\def\eqref#1{equation~\ref{#1}}

\def\1{\bm{1}}

\DeclareMathAlphabet{\mathsfit}{\encodingdefault}{\sfdefault}{m}{sl}
\SetMathAlphabet{\mathsfit}{bold}{\encodingdefault}{\sfdefault}{bx}{n}

\usepackage{hyperref}
\usepackage{url}

\usepackage{microtype}
\usepackage{multirow}
\usepackage{booktabs}

\usepackage[utf8]{inputenc}
\usepackage[T1]{fontenc}
\usepackage{times}
\usepackage{amsmath,amssymb,amsfonts}
\usepackage{pifont}
\usepackage{graphicx}
\usepackage{tikz}
\usetikzlibrary{arrows.meta,positioning,fit,backgrounds,calc}
\usepackage{natbib}
\usepackage{xcolor}
\usepackage{enumitem}
\usepackage{xspace}
\usepackage{tabularx}
\usepackage{adjustbox}
\usepackage{subcaption}
\usepackage{tcolorbox}
\usepackage[skip=10pt]{parskip}
\tcbuselibrary{breakable,skins,theorems} %

\usepackage{etoolbox} %

\usepackage{listings}
\usepackage[capitalize,noabbrev]{cleveref}
\crefname{section}{\S}{\S}   %
\crefformat{section}{\S#2#1#3}
\crefname{figure}{Fig.}{Fig.}
\crefname{table}{Table}{Tables}
\crefname{appendix}{App.}{}

\usepackage[textsize=tiny,disable]{todonotes}
\newcommand{\note}[4][]{\todo[author=#2,color=#3,size=\scriptsize,fancyline,caption={},#1]{#4}}
\newcommand{\clara}[2][]{\note[#1]{\textbf{Clara}}{orange!30}{#2}}
\newcommand{\gul}[2][]{\note[#1]{\textbf{Gül}}{purple!30}{#2}}

\definecolor{Burgundy}{RGB}{144,0,32}

\newcommand{\tokeval}{\textsc{TokEval}}

\newcommand{\brev}[1]{#1}
\newcommand{\rrev}[1]{#1}
\newcommand{\dealwithlater}[1]{}

\definecolor{blueframe}{HTML}{1769AA}
\definecolor{blueback}{HTML}{D6E9F7}
\definecolor{bluetext}{HTML}{124E7A}

\definecolor{greenframe}{HTML}{238B5A}
\definecolor{greenback}{HTML}{DDF2E8}
\definecolor{greentext}{HTML}{17603F}

\definecolor{orangeframe}{HTML}{D97706}
\definecolor{orangeback}{HTML}{FFF3E3}
\definecolor{orangetext}{HTML}{9A4F00}

\definecolor{resultframe}{HTML}{7968A5} %
\definecolor{resultback}{HTML}{F0EDF6}  %
\definecolor{resulttext}{HTML}{574581}  %

\tcbset{
  rqcolor/.is choice,
  rqcolor/blue/.style={colframe=blueframe, colback=blueback, coltitle=bluetext, titlecolor=bluetext},
  rqcolor/green/.style={colframe=greenframe, colback=greenback, coltitle=greentext, titlecolor=greentext},
  rqcolor/orange/.style={colframe=orangeframe, colback=orangeback, coltitle=orangetext, titlecolor=orangetext},
  rqcolor/purple/.style={colframe=resultframe, colback=resultback, coltitle=resulttext, titlecolor=resulttext},
  rqcolor/.default=blue,
  titlecolor/.store in=\titlecolor,
  answered in/.store in=\tcbrqanswered,
  answered in=,
}

\newtcbtheorem[
  crefname={RQ}{RQs},
  Crefname={RQ}{RQs},
]{researchq}{RQ.}{
  enhanced, breakable,
  theorem style=plain,
  before title={(},
  after title={)},
  terminator sign={},
  before upper={},
  before lower={%
    \ifx\tcbtitletext\@empty\else\tcbtitle\ \fi
    \ifdefempty{\tcbrqanswered}{}{%
      \par\vspace{2pt}\footnotesize\hyperref[\tcbrqanswered]{$\to$~\S\ref*{\tcbrqanswered}}%
    }%
  },
  boxrule=1pt,
  rqcolor=blue,
  fonttitle=\itshape,
  lower separated=false,
  boxsep=1.5pt, left=4pt, right=4pt, top=2pt, bottom=2pt,
  before skip=1ex, after skip=1ex,
  sidebyside, sidebyside align=center, sidebyside gap=1em, righthand width=3em,
}{rq}

\newtcolorbox[auto counter]{resultbox}[1][]{
  enhanced, breakable,
  rqcolor=purple,
  boxrule=0.6pt,                    %
  arc=3pt, outer arc=3pt,           %
  boxsep=1pt, left=4pt, right=4pt, top=2pt, bottom=2pt,
  before skip=0.6ex, after skip=1ex,
  fontupper=\small,
  before upper={\textbf{\color{\titlecolor}{Finding~\thetcbcounter.}}\enspace},
  #1
}
\crefname{tcb@cnt@resultbox}{Finding}{Findings}
\Crefname{tcb@cnt@resultbox}{Finding}{Findings}

\title{Language-Specific Effects of Tokenizer Choice in Multilingual Language Models}

\author{Clara Meister$^\epsilon$, G\"ul Sena Alt{\i}nta{\c s}$^{\tau,\epsilon}$, Antoine Bosselut$^\epsilon$\\
$^\epsilon$ EPFL,$^\tau$ University of Toronto, Vector Institute\\\texttt{clara.meister@epfl.ch}}

\iclrfinalcopy %
\begin{document}
\maketitle

\begin{abstract}
Tokenizer choice affects multilingual language modeling, but vocabulary capacity is finite and vocabulary size is often constrained: improving representation for some languages often comes at the expense of others.
We therefore ask whether tokenizer choice matters equally across languages, a question that results from the current literature leave unanswered. 
To this end, we train 123 language models spanning 54 tokenizers.  In the main comparison, architecture, training corpus, training-token budget, and optimization are held fixed, so the models differ only in their tokenizer. 
\clara{ how its different than "before"?}
We find that tokenizer choice matters more for languages with less language-model training data: across the 54 tokenizers, the standard deviation of a language's bits-per-byte (BPB) increases as its model training-data share decreases (Spearman $\rho=-0.52$ over the 31 trained languages and $-0.69$ over the 28 written with word boundaries).  
Leaving a language out of tokenizer training raises its BPB in every language we study, and the penalty tends to be larger for languages with less language-model training data. 
Giving lower-resource languages a larger share of tokenizer-training data, however, does not unconditionally help those languages: both equal weighting and an allocation inverting the shares with respect to the language model training data increase their BPB, particularly when language-model training repeats data. 
Finally, which intrinsic tokenizer properties are associated with better BPB differs across languages, providing further evidence that what makes a good tokenizer depends on the language.  
We find that the metrics quantifying these properties can be successfully used to predict downstream models' pairwise BPB rankings, suggesting a practical strategy for screening tokenizer candidates before training language models.\clara{gives motivation to tailor for these languages}
\begin{center}
\includegraphics[height=2.5ex]{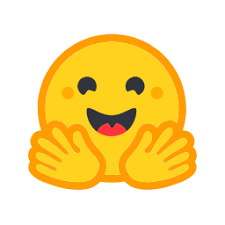} \textbf{Models \& Tokenizers: } \hyperlink{https://huggingface.co/cmeister/tokenizer-lm-ablations/}{cmeister/tokenizer-lm-ablations} \qquad \\
\includegraphics[height=2.5ex]{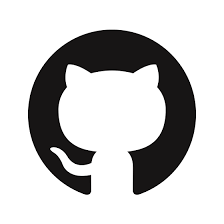} \textbf{Code: } \hyperlink{https://github.com/cimeister/tokenizer-lm-ablations}{cimeister/tokenizer-lm-ablations}
\end{center}
\end{abstract}

\section{Introduction}
\label{sec:intro}

Multilingual language models serve dozens of languages and scripts with one tokenizer that is fixed before training. %
How that tokenizer segments each language has practical consequences: prior work has raised concerns that some languages are systematically disadvantaged by current tokenizers~\citep{petrov-etal-2023-language,arnett2026Inequities}, including via large differences in API costs~\citep{ahia-etal-2023-languages}.
Controlled studies further show that tokenizer choice can substantially affect downstream model behavior when the rest of language-model training is held fixed~\citep{ali-etal-2024-tokenizer,singh-strouse-2024-tokenization,altintas2026toksuite,lesci-etal-2025-causal}.
These findings have motivated a range of modifications intended to improve multilingual tokenization~\citep{foroutan-meister-et-al-2025-parity-aware-bpe,land-arnett-2025-bpe-script,ehsaneddin2025morphbpe,jabbar2024morphpiece}.

Whether these modifications help, and for which languages, is hard to establish from existing evidence because most results are aggregated too coarsely.
Vocabulary capacity, measured in the number of unique tokens it includes, is shared and finite in a multilingual tokenizer: a change that improves the representation of one language often reduces the representation available to others.
An outcome averaged over languages can therefore change little in aggregate while individual languages' outcomes change in opposite directions.
The question of whether a tokenizer is a good choice must be posed for each language separately, and the literature lacks an account at that level: how much tokenizer choice changes a single language's performance, which languages are affected most, whether a choice that helps one language also helps others, and which tokenizer properties account for the differences.

Our work closes this gap with a large empirical study. We train 123 language models across 54 tokenizers. The 54 tokenizers vary along four design axes: pretokenization, tokenization algorithm, vocabulary encoding, and tokenizer-training data. 
The main comparison holds architecture, training corpus, training-token budget, and optimization fixed, isolating tokenizer effects on language modeling performance. 
We measure every outcome separately for each of the 31 languages of the language-model training mixture, and perform careful statistical analyses of the observed trends. With these analyses, we are able to answer four research questions (\cref{sec:rq1,sec:rq2,sec:rq3,sec:rq4}), each motivated by the result of the previous one. Their main findings are summarized below.

\begin{itemize}[noitemsep,leftmargin=15pt,topsep=-7pt]
    \item \textbf{Tokenizer choice affects every language.} Changing the tokenizer changes every language's bits-per-byte (BPB) by more than the variation across model seeds; tokenizers do not rank consistently across languages (mean pairwise Kendall $\tau = 0.418$ between the per-language rankings, \cref{sec:rq1}).
    \item \textbf{Languages with less language-model training data are the most sensitive to tokenizer choice.} The \emph{spread} of a language's BPB across the 54 tokenizers rises as its share of the language-model training data falls (Spearman $\rho=-0.52$ ); excluding a language from tokenizer training likewise raises BPB more as its share of the language-model training data falls\clara{get number} (\cref{sec:rq2}). %
    \item \textbf{Reallocating tokenizer-training data toward those languages \emph{raises} their BPB under our language model training setup.} Equal weighting and an allocation inverse to the language-model training shares both raise the BPB of the lower-resource languages (more so when language-model training repeats data), suggesting additional vocabulary for underrepresented languages is not unconditionally helpful (\cref{sec:rq3}). %
    \item \textbf{Intrinsic metrics measured per language can screen tokenizers without training a language model.} Which intrinsic tokenizer metrics are associated with lower BPB differs by language; measured per language, nine metrics order held-out tokenizer pairs by the trained models' mean BPB with $0.89$ accuracy (\cref{sec:rq4}).
\end{itemize}
From our results, we conclude that tokenizer selection is most consequential for the low-resource languages, that per-language intrinsic metrics can screen choices before any model training, and that tailoring to low-resource languages costs the high-resource languages little. %
We derive practical and actionable recommendations for multilingual tokenizer evaluation and selection (\cref{sec:recommendations}).  
To the best of our knowledge, this is the largest controlled study of tokenizer effects on multilingual language modeling performance.

\paragraph{Related studies.} Prior works have investigated how tokenizer choice changes downstream language modeling performance, spanning arithmetic \citep{singh-strouse-2024-tokenization}, code \citep{daganGettingMostOut2024,tokeval} and multilingual and general benchmarks \citep{ali-etal-2024-tokenizer,altintas2026toksuite,rust-etal-2021-good}. 
Several do this in a controlled fashion: \citet{ali-etal-2024-tokenizer} vary the algorithm and vocabulary size, \citet{goldman2024unpacking} vary the amount of tokenizer-training data, and \citet{altintas2026toksuite} compare 14 off-the-shelf tokenizers.
Others instead target a specific mechanism, such as downstream language plasticity~\citep{abagyan2025tokenizerruleallemergent} or causal estimation of the effect of a single vocabulary item~\citep{lesci-etal-2025-causal}. 
While they establish the importance of tokenizer choice, a unified account of when and how much it matters---particularly on a per-language basis---is missing. 
Each evaluates English and at most four other languages, and they do not cover the breadth of training algorithms, pretokenization and normalization strategies, text encoding and data composition in an isolated fashion. 

In parallel, mechanisms behind these differences have been studied through intrinsic metrics that could be computed from the tokenizer alone without model training.
\citet{galle-2019-investigating} and \citet{goldman2024unpacking} find that compression rate correlates with downstream performance, whereas \citet{schmidt-etal-2024-tokenization}, \citet{lester2024compression}, and \citet{lotz-etal-2025-beyond} show that compression rate alone is an incomplete predictor of downstream model quality.
Proposed alternatives include information-theoretic measures based on entropies, such as R\'enyi efficiency~\citep{zouhar-etal-2023-tokenization} and capacity utilization~\citep{erdogan2026infoTheoreticTok}, and linguistically grounded measures such as morphological alignment~\citep{arnett-bergen-2025-morphscore,arnett-etal-2025-morphscore-v2}.
\citet{tokeval} found strong correlation with information-theoretic measures, while also hinting that different intrinsics may matter more for different tasks. 
Alternatives have been proposed with varying degrees of success, but the utility of these metrics on a per-language basis has not been established yet. 

\section{Experimental Setup}
\label{sec:setup}
We train a large panel of language models that differ only in their tokenizer and evaluate them intrinsically and on downstream tasks across languages. We describe the tokenizer panel and models we train (\cref{sec:setup:training}), our evaluation metrics (\cref{sec:setup:eval}) and the statistical protocol behind our analyses (\cref{sec:setup:analysis}).

\subsection{Tokenizer and Model Training}
\label{sec:setup:training}
\paragraph{Tokenizers.}
We compare 54 tokenizers covering diverse practices from multilingual tokenization literature. We train 52 of them on mixtures of the same multilingual corpus. The other two are off-the-shelf tokenizers from the open-source LLMs Mistral-Nemo and LLaMA-3, with vocabulary sizes comparable to our panel.  
We outline our tokenizer design axes below, with more detail in \cref{app:grid}. 
\begin{itemize}[noitemsep,leftmargin=15pt,topsep=-7pt]
    \item \emph{Training algorithm.} We consider BPE~\citep{sennrich-etal-2016-neural}, UnigramLM~\citep{kudo-2018-subword}, MinGram \citep{land2026mingramminimalistunigramtokenizer}, SuperBPE~\citep{liu-etal-2025-superbpe}, parity-aware BPE~\citep{foroutan-meister-et-al-2025-parity-aware-bpe}, and combinations of the latter two.

\item \emph{Pretokenization and normalization.} 
Pretokenization determines which consecutive characters can form a token and how scripts, numbers, punctuation, and whitespace are handled (all detailed in \cref{app:pretok}). We also vary normalization across NFC, NFD, and no Unicode normalization.

\item \emph{Text encoding.} Most tokenizers operate on raw UTF-8 bytes; three instead use an encoding based on Unicode script classes \citep{land-arnett-2025-bpe-script}, proposed to reduce the byte-level penalty that scripts used by low-resource languages incur. 

\item \emph{Training data.} Our baseline \emph{balanced} tokenizer training dataset  consists of a balanced 10 GB mixture across different text domains. Concretely, its sampling weights are approximately $37\%$ English (FineWeb-Edu \citep{penedo2024fineweb}), $33\%$ multilingual across 30 languages (quality-filtered FineWeb2 \citep{penedo2025fineweb2pipelinescale} using the filtering protocol of Apertus~1.0 \citep{apertus2025}), $15\%$ math (FineMath \citep{finemath}), and $15\%$ code (StarCoderData \citep{li2023starcoder}). The 31 natural languages span 11 writing systems and 11\clara{double check this} language families. 
For six of the tokenizers, the multilingual data comes from a sample of FineWeb2's full language pool.\clara{give more info} 
We perform targeted changes to this dataset composition, ablating language coverage and per-language sampling weights, following the range of practices used to build tokenizers for existing multilingual LLMs.\clara{maybe need citation here} 
\end{itemize}

Each of these four axes has been proposed as a determinant of multilingual quality: the algorithm and the pretokenization determine which byte sequences can become vocabulary entries, the text encoding determines the per-character byte cost of each script, and the training data determines which languages' byte sequences are frequent enough to enter the vocabulary. 
Three tokenizers have known construction defects that are common artifacts of tokenizer training (a missing part of the byte-level initial alphabet, or a dropped lone space). 
We analyze the impact of the byte-alphabet defect in \cref{app:pretok-ablation} by comparing a defective tokenizer with its defect-free rebuild. 

\paragraph{Language models.}
\label{sec:training-details}
For each tokenizer, we train a corresponding 1.27\,B-parameter language model, fixing architecture, training data, and optimization across all runs following the nanochat d24 configuration~\citep{karpathy2025nanochat}.  We follow nanochat's scaling law recommendations, training on 9.2\,B tokens; \cref{app:robustness-design} shows that model rankings are robust to this design choice. 
We use the same corpora as for tokenizer training, following a composition close to that of the \emph{balanced} mixture. Unless otherwise noted for ablations, the model training mixture is held constant. {Throughout, we write \emph{LM share} for a language's share of the language-model training data, measured in bytes, and \emph{tokenizer share} for its share of the tokenizer-training data. The two need not coincide, either in our setup or in practical model development.} 
Because the number of training steps is held approximately constant, the amount of underlying text seen by each model may differ, as different tokenizers will have different compression rates. We refer to the underlying text seen by the model as its \emph{training bytes}.\clara{check this term is always used} 
Details on model architecture, optimization and training data are provided in \cref{app:training,app:data-detail}. To check whether tokenizer rankings hold beyond our main scale, models are trained for a subset of tokenizers at three smaller scales: d8, d12, d16 (\cref{tab:proxy-models}). We include these models only in the ranking analysis in \cref{sec:rq4}.

\subsection{Evaluation Metrics}
\label{sec:setup:eval}

\paragraph{Per-language outcomes.}
Our main outcome is per-language bits-per-byte (BPB)~\citep{gao-2021-the-pile}, a tokenizer-agnostic evaluation for language-modeling performance. 
Computation details are given in \cref{app:methods}\clara{add more details here and add to recommendations that this protocol be adopted}.
We use FLORES+~\citep{nllb2022flores} as our main evaluation set, a corpus of the same sentences translated into many languages. 
The main analyses use the 31 languages of the language-model training mixture. 
Results using the other 183 languages in FLORES+ are provided in  \cref{app:full-flores}.
For statements about a single language, we use that language's FLORES+ BPB.
We report BPB as the primary outcome because it is the standard unit. For a comparison across languages, however, bits per sentence on parallel text is arguably a fairer comparison as it holds content fixed; we report it alongside BPB for those comparisons (\cref{app:rq1-2}).
\gul{could be simplified}

As secondary downstream outcomes, we compute per-language accuracy on MultiBLiMP~\citep{jumelet-etal-2025-multiblimp}, a multilingual minimal-pairs grammaticality benchmark covering 24 of the 31 languages, XNLI~\citep{conneau2018xnli}, a three-way natural-language-inference benchmark covering 13, and TokSuite~\citep{altintas2026toksuite} a four-way general knowledge benchmark covering 4; these separate the models beyond seed variation. XCOPA~\citep{ponti-etal-2020-xcopa} and XWinograd~\citep{tikhonov-ryabinin-2021-heads,muennighoff-etal-2023-crosslingual} stay near chance at this scale and are reported in \cref{app:multiblimp} only.\looseness=-1

\paragraph{Aggregate outcomes.}
For comparisons that require a single outcome per tokenizer, we report two aggregate measures.
The \emph{31-language FLORES+ mean} is the mean of the run's FLORES+ BPB over the 31 languages of the language-model training mixture.
The \emph{Validation BPB} is BPB on held-out text from the language-model training mixture.
Thus, the 31-language FLORES+ mean weights all 31 languages equally and evaluates on text outside the training distribution, whereas Validation BPB reflects the unequal language proportions of the language-model training distribution. 

\paragraph{Intrinsic tokenizer metrics.}
We compute nine intrinsic tokenizer metrics with the \tokeval{} library~\citep{tokeval}.  
These metrics serve as explanatory and predictive variables in \cref{sec:metrics}. 
The metrics require no language-model training, but all the metrics we consider are computed with respect to a (user-defined) text\footnote{Some intrinsic tokenizer metrics are computable from the vocabulary alone, such as the mean token string length, but none are amongst the set that we consider.}.
Unless otherwise stated, we consider each metric on a per-language basis, i.e., computed on text from a specified language. 
\Cref{app:metrics} defines the metrics, and \cref{fig:metric-tiles} summarizes them and presents the measured coefficients of their within-language effects.

\subsection{Statistical Protocol}
\label{sec:setup:analysis}

To determine whether differences between runs exceed training noise, we estimate seed variation.
The \emph{seed variation} of a quantity is the standard deviation of that quantity across runs that differ only in the random seed. We estimate it from seed replicates of five tokenizers and pool the per-tokenizer estimates, weighting by their degrees of freedom.
For analyses of individual languages, we call a difference between two runs \emph{larger than seed variation} when its magnitude exceeds the pooled seed standard deviation for a single language's FLORES+ BPB.
For comparisons on Validation BPB and the 31-language FLORES+ mean, we use a stricter reference threshold: we call a difference \emph{larger than run-to-run variation} when its magnitude is approximately outside the 95\% range expected from two runs that differ only in random seed, i.e., when it is at least $\sim$twice the estimated standard deviation\clara{Is this one not pooled?} of the difference between two such runs. 
Exact protocol and values can be found in \cref{app:methods}. 
All $p$-values use Benjamini--Hochberg false-discovery-rate correction within the relevant set of tests. In plots and tables, $^{*}$, $^{**}$, and $^{***}$ indicates $p < 0.05$, $p < 0.01$ and $p < 0.001$, respectively.\looseness-1

We primarily report results per tokenizer, but for the sake of robustness, we additionally assess results on a per-tokenizer-family basis.
Concretely, results from the 54 tokenizers are not fully independent: some tokenizers differ only in a minor training hyperparameter or preset, such as whether the pretokenizer regex allows apostrophes to trail letters. 
This breaks independence assumptions in several of the analysis methods we use. 
We therefore additionally present results when grouping tokenizers that differ only in such minor hyperparameters into the same \emph{family}, and we rely primarily on cluster-robust standard errors, clustered at the tokenizer-family level.
Full details of the family stratification are given in \cref{app:grid} and pooled standard error computation details in \cref{app:methods}.

\section{How much does tokenizer choice matter across languages? {\normalsize(RQ1)}}
\label{sec:magnitude}\label{sec:rq1}
\begin{figure}[t]
\centering
\includegraphics[width=0.8\textwidth]{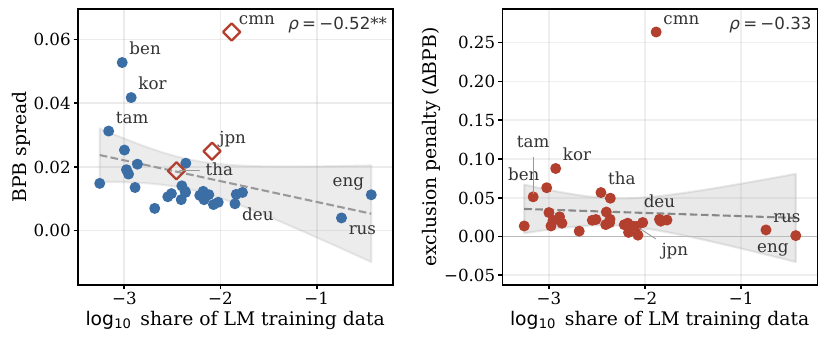}
\caption{\textbf{Left}: the standard deviation of a language's FLORES+ BPB across the 54 tokenizers (BPB spread) vs. the language's \rrev{log LM share} (Spearman $\rho = \rrev{-0.52}$, $p = \rrev{0.003}$). Open diamonds mark the three languages written without word boundaries (Chinese, Japanese, and Thai).
\textbf{Right}: the exclusion penalty (defined in the text) against the same axis (Spearman $\rho = \rrev{-0.33}$, $p = \rrev{0.068}$). The horizontal line marks a penalty of zero.
Dashed lines are ordinary-least-squares fits, and the grey regions are 95\% confidence regions for the fitted mean.}
\label{fig:spread-coverage}
\end{figure}

\begin{resultbox}[label=res:rq1]
Changing the tokenizer changes every language's FLORES+ BPB and most covered languages'  MultiBLiMP accuracy by more than that language's seed variation, and the performance of the 54 tokenizers ranks differently across languages; a tokenizer chosen on an aggregate outcome is therefore a compromise across languages rather than the best tokenizer for each of them.
\end{resultbox}
\subsection{Method}
We measure each outcome per language for each of the 54 tokenizer--model pairs. For each language, we compare the spread of the outcome across the 54 tokenizers with the seed variation of \cref{sec:setup:analysis}. We decompose the variance of the per-(tokenizer, language) measurements into a between-language and a within-language part. We measure the agreement between two languages' tokenizer rankings as the Kendall $\tau$ between the two orderings ($1$ for identical, $0$ for unrelated) and report its mean over the language pairs.
\subsection{Results}

\paragraph{Language identity accounts for nearly all of the variance in the per-(tokenizer, language) BPB measurements, so the tokenizer effect is visible only within a language.}
Language identity accounts for $99.4\%$ of the variance in per-(tokenizer, language) BPB, so a tokenizer changes only the within-language remainder.
That remainder is largest for Chinese, with a BPB standard deviation across the 54 tokenizers of $0.062$, followed by Bengali and Korean.
Every analysis that follows therefore compares tokenizers within a language.

\paragraph{The spread across tokenizers is greater than the seed variation for every language on FLORES+ BPB and for all but three on MultiBLiMP, so there is a measurable tokenizer effect for every language, not only the low-resource ones.}
The standard deviation in per-language FLORES+ BPB across the 54  tokenizers is greater than the language's seed variation for all 31 languages; on MultiBLiMP accuracy, this seed variation is exceeded for 21 of the 24 languages that MultiBLiMP covers. 
For 16 of the 24 languages, however, the seed variation is below the sampling uncertainty from the benchmark's finite number of minimal pairs, so MultiBLiMP supports fewer per-language comparisons than BPB (\cref{app:multiblimp}).
\Cref{fig:magnitude} visualizes this across languages.

\paragraph{Per-language tokenizer rankings agree only partially across languages, so no single tokenizer is best for all of them.} 
The mean pairwise Kendall $\tau$ between the per-language tokenizer rankings is $0.418$. The ranking agreement is lowest for Ukrainian ($\rrev{0.099}$).
The general disagreement is not specific to the FLORES+ text, although Ukrainian's extreme value suggests that there may be some artifacts in this particular text. 
At this level of agreement, 16 of the 54 tokenizers rank first for at least one language, i.e., none ranks first for every language.

\paragraph{Caveats.}
While within-language comparisons are invariant to the outcome's unit, cross-language comparisons of the spread depend on it, and the most sensitive language changes between BPB and bits per sentence (\cref{app:rq1-2}). The seed variation is estimated from model replicates of just five of the tokenizers (\cref{app:methods}).

In the next section, we ask which languages are affected the most by tokenizer choice, and whether a language's sensitivity is predictable from its share of the training data.

\section{Which Languages Are Most Tokenizer-Sensitive? {\normalsize(RQ2)}}
\label{sec:sensitivity}\label{sec:rq2}

\begin{resultbox}[label=res:rq2,rqcolor=blue]
Lower-resource languages, those with a smaller LM share, are more sensitive to tokenizer choice under two measures: First, excluding a language from tokenizer training raises its BPB for every language, and more so when it has a smaller share in the language-model training data. Second, each language's BPB spread across 54 tokenizers also falls as its LM share rises. We find writing system and resource-level effects are confounded (\cref{app:rq1-2}).  
\end{resultbox}

\subsection{Method}
We use two per-language measures of sensitivity: the spread from \cref{sec:rq1} and the exclusion penalty. For the latter, we train four ``coverage tokenizers,'' tokenizers that differ only in which languages their tokenizer-training data include (English only, the 6 or 21 languages with the largest LM shares, and all 31 languages). A language's penalty is then its mean FLORES+ BPB over the coverage tokenizers that exclude it minus its mean over those that include it. We correlate both measures with the logarithm of the language's LM share (Spearman). \Cref{app:rq1-2} tests whether either relationship is accounted for by the writing system, and \cref{app:robustness-design} whether it is accounted for by the differences in bytes per character between languages.

\subsection{Results}

\paragraph{A language's BPB varies less across tokenizers as its LM share rises, so the languages with the least language-model training data are most sensitive to tokenizer choice.}  
\Cref{fig:spread-coverage} (left) shows tokenizer performance spread (standard deviation of FLORES+ BPB across the 54 tokenizers) as a function of log {LM share}. The two are negatively correlated (Spearman $\rho = -0.52$), significant. 
The correlation is negative and significant under every variant we test: excluding the three languages written without word boundaries, aggregating the spread by tokenizer family rather than by individual tokenizers, measuring the outcome in bits per sentence, and dividing the spread by the language's mean BPB (\cref{app:rq1-2}).
Russian and German, two of the highest-resource languages in the language-model training data are clear examples: after dividing each language's spread by that language's mean FLORES+ BPB across the 54 tokenizers, they have the smallest and third-smallest coefficient of variation in FLORES+ BPB across tokenizers.
Chinese is the main exception, despite having the sixth-largest share, it has the largest spread and the largest exclusion penalty.
We find that this ranking is sensitive to measurement strategy, and under the alternative approach in \cref{app:rq1-2}, Bengali and Tamil, the two lowest resource languages, have a larger spread than Chinese.
\paragraph{Excluding a language from the tokenizer-training data raises its FLORES+ BPB for every language; the penalty is negatively correlated with the language's LM share.}
No coverage tokenizer excludes English, so English's excluded mean comes from a tokenizer that is part of a separate set of ablations ( where the tokenizer-training data contains no English web text; \cref{app:inversion}).
The exclusion penalty is positive for all 31 languages\footnote{This partly reflects a general negative effect of narrower tokenizer training data coverage: tokenizers that exclude more languages have higher mean FLORES+ BPB across all 31 languages \rrev{(\cref{tab:exclusion-general-difference} in \cref{app:inversion})}.
},
with \rrev{Chinese ($0.264$) and Korean ($0.088$)} having the largest penalties.
\Cref{fig:spread-coverage} (right) plots exclusion penalty against log LM share; the correlation is negative (Spearman $\rho=-0.33$), but not significant, in BPB and in bits per sentence alike (\cref{app:rq1-2}). 
The relationship is significant, though, under either of two controls: restricting the analysis to the 15 languages excluded by at least two coverage tokenizers, and adjusting the share for the language's bytes per character (\cref{app:robustness-design}). 
We perform further analyses on potential confounds for this result in \cref{app:robustness-design}.\looseness=-1

\paragraph{One tokenizer can be close to the best choice for every sensitive language at once, at no measurable cost to the languages least affected by tokenizer choice.}
\Cref{sec:rq1} shows that no single tokenizer ranks first for every language. For the high-resource languages a lower rank corresponds to a small difference in BPB, because their BPB spread across the tokenizers is the smallest. For the 10 languages with the smallest \rrev{LM shares}, where the spread is largest, the rankings of the tokenizers largely agree: the tokenizer that ranks first on the 31-language FLORES+ mean ranks no lower than 11th of 54 on each of those 10 llanguages (BPE Clean plus3 NFC on the balanced data, \cref{tab:grid}). As further evidence of this, in the coverage tokenizers above, going from the tokenizer trained on the six high-resource languages to the tokenizer trained on all 31 (\cref{tab:select}) lowers FLORES+ BPB for 30 of the 31 languages, while English BPB and validation BPB changed by less than their seed variation.

\paragraph{Caveats.}
Resource level and writing system are confounded in our 31 language set (\cref{app:rq1-2}). The penalty--share correlation is not significant without controls; it is significant within the 15 languages excluded by two coverage tokenizers and after adjusting for bytes per character (\cref{app:robustness-design}). Part of each penalty is a general overall rise in model BPB under narrower tokenizer language coverage (\cref{tab:exclusion-general-difference}).
\gul{this sounds unnatural}

Both measures of this section are associations with the LM share, not interventions on it: languages that differ in share differ in other properties too. In \cref{sec:rq3} we intervene: we change one design choice at a time, among them the per-language tokenizer shares, and measure each change's per-language effect.

\section{Do design choices change per-language outcomes? {\normalsize(RQ3)}}
\label{sec:design}\label{sec:rq3}
\begin{resultbox}[label=res:rq3,rqcolor=green]
Training the tokenizer on only English raises FLORES+ BPB \rrev{for nearly every language}, having more of an impact than any other single configuration change we tested. We test two reassignments of the tokenizer training data sampling weights that raise lower-resource languages' tokenizer shares. Under both, the mean FLORES+ BPB of these languages is \emph{higher}, not lower.
\end{resultbox}

\subsection{Method}
We change one design choice at a time and compare the resulting tokenizer--model pairs in two ways. First, we train 25 pairs that each differ from a matched baseline in a single configuration choice: pretokenization, algorithm, normalization, vocabulary encoding, or tokenizer-training data (\cref{app:pretok-ablation}). 
For each pair, we report the change in the 31-language FLORES+ mean and in validation BPB against the run-to-run thresholds of \cref{sec:setup:analysis}. Second, keeping the total budget fixed, we reallocate the per-language tokenizer shares (equal shares, two permuted allocations, an inverted allocation, and a variant without English web text) against a fixed reference pair. We then correlate each language's BPB change with its share. Because changing the tokenizer while keeping a fixed token budget also changes the number of training bytes, we retrain selected pairs matched on training bytes or on per-language token counts, and we repeat the reallocation comparison without any data repetition (\cref{app:inversion}). 

\subsection{Results}

\paragraph{Restricting the tokenizer-training data to English changes BPB more than any other single configuration change.}
\Cref{tab:select} (\cref{app:pretok-ablation}) lists the 25 controlled pairs with each pair's difference on the 31-language FLORES+ mean and on validation BPB.
The largest of the 25 differences comes from restricting the tokenizer-training data to English ($\rrev{+0.0386}$ in the 31-language FLORES+ mean).
Under the coverage tokenizers in the previous section, where we restrict tokenizer-training data to just the 6 high-resource or 21 high- and mid-resource languages, both result in smaller performance degradations.
Of the 25 design choices ablated, 18  show a difference larger than run-to-run variation on the 31-language FLORES+ mean (16 on validation BPB). 
We discuss the others in \cref{app:pretok-ablation}.

\paragraph{Raising the tokenizer-training weight of the low-resource languages to an equal share raises their BPB instead of lowering it.} 
For five tokenizer configurations in our main set, we train an \emph{equal weight} version, where all 30 FineWeb2 languages are sampled at the same weight instead of weights proportional to each language's amount of FineWeb2 data; English, drawn from FineWeb-Edu, keeps its baseline sampling weight ($0.35$). 
The exact protocol is described in \cref{app:inversion}. 
We compare pairwise performance; values are given in \cref{tab:equalweight}.\footnote{For any given pair, both language models are trained on the same number of tokens, but because different tokenizers will segment the underlying text differently, the training bytes across the models differ.}
In all five pairs, the average FLORES+ BPB over the 15 languages with the smallest proportional tokenizer-training weights is higher under the equal weighting scheme. Further, the 31-language FLORES+ mean is also higher. For four of the tokenizers, this rise is larger than run-to-run variation.

\begin{figure}[t]
\centering
\includegraphics[width=0.8\textwidth]{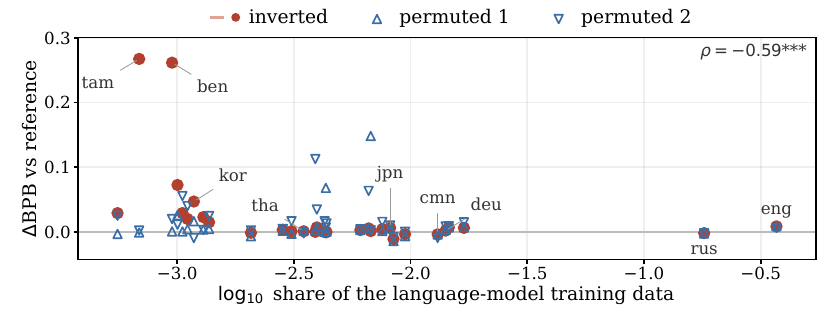}
\caption{Per-language change in FLORES+ BPB relative to reference (the model trained with the GPT-4o BPE reference tokenizer) for the 31 languages of the training mixture, as a function of the language's \rrev{log LM share}.
Filled circles: inverted-allocation tokenizer, in which languages with the smallest nominal sampling weights in the language-model training data receive the largest tokenizer shares. 
Open triangles: the two permuted allocation tokenizers, in which the tokenizer shares are assigned randomly.\looseness=-1   
}
\label{fig:inversion}
\end{figure}

\paragraph{Every change to the languages' tokenizer shares raised the 31-language FLORES+ mean, and when the languages with the least language-model training data received the largest shares, FLORES+ BPB rose most for the languages with smaller shares of language-model training data.}
Under the equal weight scheme, the shares of all languages change at once, so we cannot observe what happens to a single language whose tokenizer share rises.
We therefore train four additional tokenizers ablating just the languages' tokenizer shares. 
We use  the balanced GPT-4o BPE tokenizer as our reference point, referring to it as  the \emph{reference tokenizer} and to its corresponding language model as the \emph{reference model}. 
In what we call the \emph{inverted-allocation tokenizer}, the languages with the smallest nominal sampling weights in the language-model training data receive the largest tokenizer shares (Tamil receives the largest).  
The other three configurations are one without English web text and two with random permutations of the same shares computed for the inverted-allocation tokenizer (permuted allocations 1 and 2). 
We train the models of the four additional tokenizers for more steps than the reference model, so that every model is trained on approximately the same total amount of training bytes. 
Results are similar  when we instead keep the number of training tokens matched (\cref{app:inversion}). 
Under each of the four targeted configurations, the 31-language FLORES+ mean is higher than the reference model's by more than run-to-run variation (largest rise: $+0.0256$, inverted-allocation tokenizer). 
Under the inverted-allocation tokenizer,
the change in FLORES+ BPB is larger for languages with a smaller LM share (Spearman correlation $-0.59$; \cref{fig:inversion}). 
Tamil, which receives the largest tokenizer share, has the largest rise in FLORES+ BPB of the 31 languages, $+0.267$. 
The two permuted allocations give us a reference point: the same correlation is $\rrev{+0.08}$ and $\rrev{-0.28}$, respectively, neither significant (\cref{tab:inversion-arms} in \cref{app:inversion}). 
Together with the equal-weighting results, this suggests that allocating vocabulary entries to a language is not always helpful to that language (and is perhaps actually harmful). We suspect this may be a by-product of insufficient language-model training data in the respective languages to fit these added entries' embeddings.

\setlength{\belowcaptionskip}{-4pt}
\clara{I suppose the bar is just not visible bc seed variation is small. Double-check}

Sampling the tokenizer-training data from FineWeb2's full language pool instead of the 31-language mixture also raises BPB: against matched counterparts, the full-pool tokenizers' mean FLORES+ BPB is higher for the 31 trained languages in three of four pairs, and higher for the 183 evaluated languages outside the language-model training mixture in all four pairs, by $+0.20$ to $+0.42$ (\cref{app:sec:inversion:compared}).

\paragraph{When language-model training data is not repeated, Tamil's FLORES+ BPB rises about half as much under the tokenizer with inverted data allocation, and FLORES+ BPB  no longer rises most for the languages with smaller shares of language-model training data. }\clara{wording of 'inverted data allocation'}
Every model discussed so far makes two to three passes over the language-model training mixture. 
This is arguably the more common setting for language model training on lower-resource languages, where data must be repeated \cite{muennighoff2023scaling}. 
To test the impact of this design choice, we train models for the reference tokenizer and the inverted-allocation tokenizer on a larger language-model training mixture of the same composition, in which no document is processed twice, i.e., single-pass training.  
In this regime, Tamil's FLORES+ BPB under a model trained with the inverted-allocation tokenizer is still significantly higher than a model trained with the reference tokenizer ($+0.14$ vs. $\rrev{+0.267}$ with repeated data). 
However, in single-pass training, the correlation between a language's change in FLORES+ BPB and its LM share is positive ($+0.34$) and not significant. 
The association between a language's LM share and its rise in FLORES+ BPB under the inverted-allocation tokenizer therefore appears only when language-model training repeats data.

\paragraph{Caveats.}
At a fixed token budget, a tokenizer change also changes the training bytes; retrains matched on training bytes reverse one of the 25 comparisons (\cref{app:pretok-ablation}). The share--rise correlation under the inverted allocation appears only under our canonical model training setup, when language-model training repeats data (\cref{app:inversion}). 

A better tokenizer for the sensitive languages comes from choosing among candidate configurations. In the next section, we ask whether there are efficient ways to identify such tokenizers, i.e., whether we can predict the per-language outcomes well enough to replace most LM training runs with a screening step. 

\section{Which tokenizer properties matter for which language? {\normalsize(RQ4)}}
\label{sec:metrics}\label{sec:rq4}
\begin{resultbox}[label=res:rq4,rqcolor=orange]
No intrinsic tokenizer metric trends with BPB the same way for every language. 
Within a language, there are consistent and significant trends;
which of these relationships holds, and how strongly, changes with the language's LM share. 
Using intrinsic metrics measured on a per-language basis, we can train pairwise tokenizer BPB rankers, whose held-out accuracy is sufficient for screening tokenizers during development.
\end{resultbox}

\emph{Disclaimer:} Every relationship between a metric and FLORES+ BPB or validation BPB in this section is a correlation across tokenizers, and not a causal claim.

\subsection{Method}\label{sec:logistic}
\paragraph{Predictive ranking.}
We train logistic pairwise rankers that predict, from the difference of two tokenizers' intrinsic-metric vectors, which of the two trained models has the lower BPB (implementation details in \cref{app:proxy-details}).
To account for dependence among closely related tokenizers, we use leave-one-family-out cross-validation: each fold holds out an entire tokenizer family, so a tokenizer is never ranked by a model trained on another member of its own family.
We evaluate pairwise accuracy, the fraction of scored tokenizer pairs whose predicted ordering matches their ordering by the trained models' BPB, over every pair with at least one tokenizer in the held-out family.\looseness=-1

\paragraph{Associations between tokenizer metrics and downstream outcomes.}
Absolute BPB differs substantially across languages, with between-language differences much larger than the tokenizer-induced differences within any one language (\cref{sec:magnitude}). 
For the analyses asking how a tokenizer metric $m$ is associated with a downstream outcome across languages and tokenizers, this artifact must be accounted for, which we do using linear mixed-effects model with per-language random effects\clara{add footnote pointing to resource}.
For example,  when FLORES+ BPB of tokenizer $t$ on language $\ell$ is the outcome, we fit:
\begin{equation}
  \text{BPB}_{t,\ell}
  =
  \beta_0
  + \beta_m \, z(m_{t,\ell})
  + b_\ell
  + \varepsilon_{t,\ell},
  \qquad
  b_\ell \sim \mathcal{N}(0,\tau^2),
  \quad
  \varepsilon_{t,\ell} \sim \mathcal{N}(0,\sigma^2),
\end{equation}
where $z(m_{t,\ell})$ standardizes the metric using the mean and standard deviation computed over all $(t,\ell)$ pairs in the analysis sample.
The random intercept $b_\ell$ accounts for language-specific BPB baselines; $\beta_m$ then gives the change in BPB associated with a one-standard-deviation increase in $m$.
To test whether this relationship varies by resource level, writing system, or morphological type, we add the corresponding interaction terms in the relevant analyses: an interaction term is the product of the metric and the language attribute, and its coefficient is the change in the metric's coefficient per one-unit change of the attribute.

\subsection{Results}
\paragraph{Measuring intrinsic metrics per language, rather than as one value per tokenizer, enables tokenizer ranking without model training that is accurate enough for screening configurations without LM training.}
We ask whether intrinsic tokenizer metrics can be used to rank tokenizers by their corresponding LM's BPB, allowing us to skip language model training for preliminary tokenizer assessment. 
We train the logistic models described in \cref{sec:logistic}, using the intrinsic metrics as features. We score their pairwise ranking accuracy: the fraction of held-out tokenizer pairs that a metric-based ranker orders the same way as the trained models' BPB. 
In each cross-validation fold, we hold out one of the 42 tokenizer families, so a held-out tokenizer is never ranked by a model fit on members of its own family.
With the nine metrics as one value per tokenizer, these classifiers order held-out tokenizer pairs according to their validation BPB at accuracy $0.700$ and according to their 31-language FLORES+ mean BPB at accuracy $0.735$. %
With the same nine metrics measured separately on each of the 31 languages, 
the accuracies rise to $0.907$ and $0.890$, respectively (\cref{tab:proxy}).
These accuracies are higher than those from ranking the tokenizers by the validation BPB of smaller models trained with them (\cref{tab:proxy}). 
Ranking by intrinsic metrics can also be done accurately on a per-language basis:
when predicting pairwise ranks of individual language's own FLORES+ BPB, the per-language ranker scores between $0.738$ (Czech) and $0.878$ (Tamil).
These results indicate that intrinsic metrics can be used for screening tokenizer choice, both at the aggregate and per-language level.

\begin{figure}[t]
\centering
\includegraphics[width=0.85\textwidth]{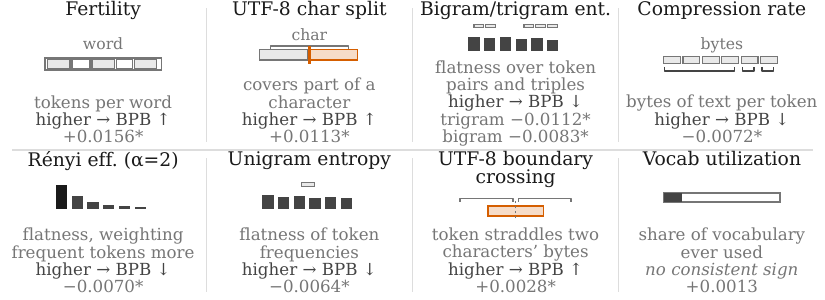}
\caption{The nine intrinsic metrics we consider and their within-language relationships with FLORES+ BPB (lower $\downarrow$ better). Each tile gives the metric's definition, the direction of the relationship, and the coefficient from the mixed effects model of \cref{sec:logistic}, in BPB per standard deviation of the metric.\looseness=-1
}
\label{fig:metric-tiles}
\end{figure}

\paragraph{Within a language, all tokenizer intrinsic metrics except vocabulary utilization are significantly related to FLORES+ BPB and MultiBLiMP accuracy.}
\label{sec:within}
We fit the linear mixed-effects models described in \cref{sec:logistic} with one intercept per language, once for each of the nine intrinsic tokenizer metrics shown \cref{fig:metric-tiles}.
Both the outcome variable and the intrinsic metric are computed on a per-language basis. 
With per-language FLORES+ BPB as the outcome and per-language intrinsic metrics measured on FLORES+, eight of the nine coefficients corresponding to the different intrinsic metrics are significant (\cref{tab:within} in \cref{app:metrics}).
Tokenizers with a higher unigram, bigram, or trigram entropy, a higher R\'enyi efficiency, or a higher compression rate on a language have a lower FLORES+ BPB on that language.
Tokenizers with a higher fertility, a higher character-split rate, or a higher boundary-crossing rate have a higher FLORES+ BPB. 
The coefficient of vocabulary utilization is not significant.
With per-language MultiBLiMP accuracy as the outcome, on the 24 languages that MultiBLiMP covers, the same eight coefficients are significant, each with the reversed sign (since higher is better for this metric).
With intrinsic metrics and BPB computed on FineWeb2 text, all nine coefficients are significant, and the coefficient of vocabulary utilization is intuitively negative (\cref{app:fineweb2}): within a language, BPB is lower for the tokenizers for which a larger fraction of the vocabulary entries occurs in that language's text.
Notably, when we adjust each run's FLORES+ BPB for its number of training bytes in \cref{app:methods:bytes}, the coefficient of compression rate is significant under only one of our four estimates of the adjustment. 
This suggests that several previous results in the literature finding that tokenizer compression rate correlates with model performance (\cref{sec:intro}) were perhaps confounded by differences in underlying training bytes across the  models tested. 

\paragraph{For single languages, different metrics are significantly correlated with FLORES+ BPB.}
For each of the 31 languages and each intrinsic metric, we compute the Spearman correlation between the metric and FLORES+ BPB across the 54 tokenizers.  
Which metrics correlate significantly with downstream performance differs greatly across languages. 
Correlation heatmaps visualize this result in \cref{app:extra-results}. 
\rrev{As a concrete example, the correlation of R\'enyi efficiency is significant for Greek ($-0.72$) and near zero for Tamil ($+0.17$).}
In \cref{app:fineweb2}, we see the results are robust to choice of evaluation corpus, and in \cref{app:methods:bytes}, we only find a significant interaction between the metric and the language's LM share for two of the nine metrics, %
suggesting these relationships are largely properties of the language rather than artifacts of the model training setup.
Collectively, we take this as evidence that different tokenizer properties are important for different languages.

\paragraph{Caveats.}
The BPB--intrinsic metric relationships are less strong once we control for the number of training bytes: within-language coefficients from the mixed-effects models shrink and about three-fifths of the significant per-language BPB--intrinsic metric correlations remain significant under the measured-training-bytes partial correlation (\cref{app:methods:bytes}). 
There may be artifacts in what each of the intrinsic metrics measures, causing spurious weakening or strengthening of the observed BPB--intrinsic metric relationships; the SuperBPE tokenizers are a measured case (\cref{app:metrics}). 

\section{Recommendations}
\label{sec:recommendations}

\rrev{Four} recommendations follow from the answers to the four questions (\cref{sec:rq1,sec:rq2,sec:rq3,sec:rq4}).%
\begin{itemize}[leftmargin=15pt,noitemsep,topsep=-5pt]
\item \emph{Per-language evaluation.} Evaluate candidate tokenizers on each language the model is meant to serve, or at least stratified by resource tiers and writing systems. Tokenizer rankings agree only partially across languages (\cref{sec:rq1}), and aggregate or English-only evaluations give little or no weight to the languages with the least language-model training data, which are the languages most sensitive to the tokenizer (\cref{sec:rq2}). %
\item \emph{Caution with reallocating tokenizer capacity toward low-resource languages.} Every reallocation we tested raised the BPB of the lower-resource languages that it was meant to favor, and every departure from proportional training data weights raised the 31-language FLORES+ mean BPB (\cref{sec:rq3}). We do not claim that such reallocations are universally harmful, as ours is a single training setting, but these results argue that targeted allocations of the vocabulary to lower-resource languages are inteventions that should be validated, not presumed beneficial.  Consequently, we recommend proportional tokenizer-training weights as the default to start from. 
\item \emph{Metric-based screening before training.} Use intrinsic metrics, measured per language, to screen candidate tokenizers before any model training as they provide good signal for tokenizer quality (\cref{sec:rq4}). Importantly, these metrics should be used as screening statistics, not training objectives.

\item \emph{Evaluation protocol.} Compute BPB against the byte count of the original text, never against the tokenizer's decoded output, whose normalization can change the byte count. Follow the same formatting protocol as used in training, e.g., prepend the BOS token. Where a parallel corpus is available, also report bits per sentence: parallel sentences hold the content fixed, so comparisons across languages do not depend on arbitrary artifacts, such as a language's bytes per character in the encoding scheme.
\end{itemize}

\section{Limitations}
Our results use one architecture (nanochat d24, decoder-only) at 1.27B parameters, one training recipe, one vocabulary size near 128K, and, for most of the 54-tokenizer set, one run per tokenizer.
Our primary outcome is language-modeling performance (BPB). Of the task benchmarks we evaluated, MultiBLiMP and XNLI separate the models: XCOPA and Belebele accuracies stay near chance at this scale (\cref{app:multiblimp}), so our conclusions concern language-modeling performance, not downstream task ability.\gul{check}
The 54 tokenizers are an observational panel rather than a randomized design: the metric analyses of \cref{sec:rq4} are correlations across this panel, and a metric that co-varies with the ablated design choices cannot be fully separated from them.
The tokenizer family partitioning is not perfect: some near-duplicate tokenizers may be considered part of different families due to how we chose to delineate families. Consequently, there may still be correlated errors that shrink our standard error estimates, causing us to identify results as significant when they are not.
Every run in our main analysis makes two to three passes over the language-model training mixture. Aggregate model rankings change little without repetition or at twice the token budget (Spearman $+0.93$ and $+0.94$ over 11 runs; \cref{app:robustness-design}), but several per-language results interact with repetition, e.g., the reallocation experiments (\cref{sec:rq3}).
Finally, the 31 languages are those of our training mixture; statements about which languages are most sensitive to tokenizer choice are statements within this set.

\section*{Reproducibility Statement}
\cref{app:grid} lists every tokenizer with its exact configuration, vocabulary size, and family assignment, and \cref{app:pretok} describes each pretokenization regex. \cref{app:training} states the full architecture, optimizer settings, and step counts, and \cref{app:data-detail} the composition of the language-model training mixture with per-language byte shares (\cref{tab:lang}). \cref{app:methods} specifies the BPB computation, the seed-variation estimates, and the clustered standard-error construction, and \cref{app:proxy-details} the ranking protocol, which is deterministic and uses no random seeds. All evaluation data are public (FLORES+, MultiBLiMP, XCOPA, XNLI, TokSuite, Wino-X, Belebele). Every number in the paper is bound, in the paper source, to a named key of a result artifact computed programmatically from the run outputs. We will release the tokenizers, model checkpoints, result artifacts, and the training, evaluation, and analysis code upon de-anonymization. %

\section*{The Use of Large Language Models}
Large language model (LLM) based tools were used in two ways in this work.
First, for writing assistance: polishing author-written text for grammar and clarity and checking terminology for consistency.
Second, as agents assisting with experiment management, for example launching training runs, monitoring them, and collecting their outputs into result files.
The research questions, experimental design, analyses, and conclusions are the authors' own.
Every number reported in this paper is bound, in the paper source, to a named key of a result artifact that was computed programmatically from the run outputs, and the authors checked these bindings against the underlying artifacts to ensure that no LLM-introduced errors (hallucinated or mistranscribed values) are present in the reported results.

\subsubsection*{Acknowledgments}
We thank Colin Raffel and Sander Land for helpful feedback on the presentation of this work, and Deniz Bayazit and Ilia Mahrooghi for feedback on an earlier version of this manuscript.
The computational resources used in this project were provided by the Swiss AI Initiative.

\bibliography{colm2026_conference}
\bibliographystyle{iclr2027_conference}

\clearpage
\appendix

\section{Experimental Details}

\subsection{Tokenizer Configurations}
\label{app:grid}

This section expands on \cref{sec:setup}, where \cref{tab:grid} lists the 54 tokenizers with their configuration and their family (\cref{sec:setup}); a tokenizer without a family code forms a single-member family.

\paragraph{Tokenizer families.}
We define tokenizer families as groups of tokenizers differing in only minor hyperparameter settings. Differences in algorithm, pretokenization regex, normalization, encoding, or training data are treated as larger design changes, and so we categorize the tokenizers differing on these axes into different families. There are two cases that can arguably be seen as exceptions: we group variants of one pretokenization regex that differ in single clauses into one family, and we also group each pair of tokenizers whose training data differ only in sampling settings into one family. 

\paragraph{Tokenizer training.}
We train tokenizers with a vocabulary near 128K on 10\,GB of text drawn from the data mixtures used for language model training (\cref{sec:data}) or their restrictions and reweightings. Most tokenizers are trained using the HuggingFace \texttt{tokenizers} library. 
The SuperBPE, parity-aware BPE, and SCRIPT tokenizers were built with separate training code, based off the codebases released by the respective authors. 
The off-the-shelf Mistral-Nemo vocabulary is $131{,}072$ (about 1K of them special tokens) and the LLaMA-3 vocabulary is $128{,}256$.

\paragraph{Defective tokenizers.}
Two implementation defects affect some panel members, these are marked with footnotes in \cref{tab:grid}.
The byte-alphabet defect (marker~$^b$, described in \cref{sec:setup}) occurs when \emph{the base alphabet of a byte-level tokenizer lacks some of the 256 byte symbols, so those bytes cannot be represented by the initial vocabulary}. Where a defect-free counterpart exists, we use the full 256-entry byte-alphabet build (marker~$^a$). For example, the english-data GPT-4o row is such a re-build, which replaced the original defective tokenizer among the 54.
The lone-space drop (marker~$^c$) is that \emph{the decoder's Metaspace convention removes a token consisting only of a space, so a lone space is lost in the decoded text}.
We retain the three marked defective tokenizers ($b$ and $c$) because no defect-free counterpart exists.

\paragraph{Variations of tokenizer-training data coverage.}
Some of the 54 tokenizers differ only in which languages their training data covers, and one further experiment varies how that data is allocated across languages; we describe both here.
The coverage sequence used in \cref{sec:sensitivity} and \cref{tab:select} is the set of GPT-4o-regex BPE tokenizers with data marked english, highres, highmid, and balanced in \cref{tab:grid}.
For the reallocation experiment (\cref{app:inversion}), we trained four additional tokenizers whose training data is allocated across languages in deliberately unusual ways.  
In the inverted-allocation tokenizer, the languages with the least language-model training data receive the largest shares of the tokenizer-training data; in the two permuted allocations, the same 31 shares are assigned in two random orders to the original balanced mixture.
We exclude these four tokenizers from the 54 tokenizers in the main analyses, as such allocations are unrealistic for tokenizers meant for practical use, and including them would skew the statistics that we compute over the main panel, such as the standard deviation of each language's BPB across tokenizers and the ranking of tokenizers within each language.

\begin{table}[!htbp]
\centering
\caption{The 54 tokenizers. Pretok: pretokenization regex (\cref{app:pretok}).
Data: tokenizer-training mixture (balanced multilingual; english-only; code-heavy;
the highres / highmid coverage subsets; all-equal reweighting; balanced-half;
multi-heavy; the full-language-pool FineWeb2 sample, fw2full; off-the-shelf
tokenizers are not trained by us). Norm.: Unicode normalizer. Family: tokenizers
sharing a code form one family under the rule of \cref{sec:setup}; codes abbreviate the identifiers of the family partition, and a blank entry
is a single-member family. Vocab: the tokenizer's vocabulary size including
special tokens, read through the evaluation loader; this is not the padded
model embedding size.
$^{a}$full 256-entry byte-alphabet rebuild (described in \cref{app:pretok-ablation}); 
$^{b}$byte-alphabet defect (described in \cref{app:pretok-ablation});
$^{c}$lone-space drop through the decoder's Metaspace convention (\cref{sec:setup}); 
$^{s}$SCRIPT text encoding (\cref{sec:setup});
$^{d}$NFD, followed by the deletion of the control characters CR, VT, FF, and NEL and the replacement of Unicode space characters by the ASCII space (\texttt{U+0020}).}
\label{tab:grid}
\small
\adjustbox{max width=\textwidth}{%
\begin{tabular}{lllllrc}
\toprule
\textbf{Tokenizer} & \textbf{Algorithm} & \textbf{Pretok} & \textbf{Data} & \textbf{Norm.} & \textbf{Vocab} & \textbf{Family} \\
\midrule
BPE GPT-4o (balanced)            & BPE & GPT-4o      & balanced          & --- & $128{,}256$ & G \\
BPE GPT-4o (balanced, file cap)  & BPE & GPT-4o      & balanced          & --- & $128{,}256$ & G \\
BPE GPT-4o NFC (balanced)        & BPE & GPT-4o      & balanced          & NFC & $128{,}256$ &   \\
BPE GPT-4o (english)$^{a}$       & BPE & GPT-4o      & english           & --- & $128{,}256$ &   \\
BPE GPT-4o (code)                & BPE & GPT-4o      & code              & --- & $128{,}256$ &   \\
BPE GPT-4o (highres)             & BPE & GPT-4o      & highres           & --- & $128{,}256$ &   \\
BPE GPT-4o (highmid)             & BPE & GPT-4o      & highmid           & --- & $128{,}256$ &   \\
BPE GPT-4o (all-equal)           & BPE & GPT-4o      & all-equal         & --- & $128{,}256$ & fam05 \\
BPE GPT-4o (all-equal, no repeat)& BPE & GPT-4o      & all-equal         & --- & $128{,}256$ & fam05 \\
BPE GPT-4o NFC (all-equal)       & BPE & GPT-4o      & all-equal         & NFC & $128{,}256$ &   \\
BPE GPT-4o NFC (fw2full)         & BPE & GPT-4o      & fw2full           & NFC & $128{,}000$ &   \\
BPE GPT-4o script-enc.\ (bal.)$^{s}$ & BPE & GPT-4o  & balanced          & NFC & $128{,}256$ &   \\
BPE Claude (balanced)            & BPE & Claude      & balanced          & --- & $128{,}256$ &   \\
BPE Claude NFC (balanced)        & BPE & Claude      & balanced          & NFC & $128{,}256$ &   \\
BPE Claude (english)$^{b}$       & BPE & Claude      & english           & --- & $128{,}256$ &   \\
BPE Claude (all-equal)           & BPE & Claude      & all-equal         & --- & $128{,}256$ &   \\
BPE Claude V2 (balanced)         & BPE & Claude V2   & balanced          & NFD$^{d}$ & $128{,}256$ &   \\
BPE GPT-2 (balanced)             & BPE & GPT-2       & balanced          & --- & $128{,}256$ &   \\
BPE Punct (balanced)             & BPE & Punct       & balanced          & --- & $128{,}256$ &   \\
BPE Punct (english)$^{b}$        & BPE & Punct       & english           & --- & $128{,}256$ &   \\
BPE Punct (all-equal)            & BPE & Punct       & all-equal         & --- & $128{,}256$ &   \\
BPE RightAlign (balanced)        & BPE & RightAlign  & balanced          & --- & $128{,}256$ &   \\
BPE RightAlign NFC (balanced)    & BPE & RightAlign  & balanced          & NFC & $128{,}256$ &   \\
BPE Whitespace (balanced)        & BPE & Whitespace  & balanced          & --- & $128{,}256$ &   \\
BPE Whitespace (multi-heavy)     & BPE & Whitespace  & multi-heavy       & --- & $128{,}256$ &   \\
BPE Clean NFC (balanced)         & BPE & Clean       & balanced          & NFC & $128{,}004$ & fam00 \\
BPE Clean no-mark NFC (bal.)     & BPE & Clean no-mark & balanced        & NFC & $128{,}000$ & fam00 \\
BPE Clean plus2 NFC (bal.)       & BPE & Clean plus2 & balanced          & NFC & $128{,}256$ & fam00 \\
BPE Clean plus3 NFC (bal.)       & BPE & Clean plus3 & balanced          & NFC & $128{,}256$ & fam00 \\
BPE Clean NFC (fw2full)          & BPE & Clean       & fw2full           & NFC & $128{,}000$ &   \\
Mistral-Nemo                     & BPE & Mistral-Nemo & off-the-shelf    & --- & $131{,}072$ &   \\
LLaMA-3                          & BPE & LLaMA-3     & off-the-shelf     & --- & $128{,}256$ &   \\
\midrule
Unigram GPT-4o (balanced)        & UnigramLM & GPT-4o     & balanced     & --- & $128{,}256$ & D \\
Unigram GPT-4o tuned (balanced)  & UnigramLM & GPT-4o     & balanced     & --- & $128{,}256$ & D \\
Unigram GPT-4o (highres)         & UnigramLM & GPT-4o     & highres      & --- & $128{,}256$ &   \\
Unigram GPT-4o (highmid)         & UnigramLM & GPT-4o     & highmid      & --- & $128{,}256$ &   \\
Unigram Claude (balanced)        & UnigramLM & Claude     & balanced     & --- & $128{,}256$ &   \\
Unigram RightAlign (balanced)    & UnigramLM & RightAlign & balanced     & --- & $128{,}256$ &   \\
Unigram Whitespace (balanced)$^{c}$ & UnigramLM & Whitespace & balanced  & --- & $128{,}256$ &   \\
\midrule
MinGram script-enc.\ (balanced)$^{s}$ & MinGram & SCRIPT   & balanced    & NFC & $128{,}256$ & fam06 \\
MinGram script-enc., split newlines$^{s}$ & MinGram & SCRIPT & balanced  & NFC & $128{,}256$ & fam06 \\
\midrule
SuperBPE GPT-4o (balanced)       & SuperBPE & GPT-4o      & balanced     & --- & $128{,}004$ &   \\
SuperBPE Clean C2 (balanced)     & SuperBPE & Clean       & balanced     & NFC & $128{,}004$ & C \\
SuperBPE Clean C3 (balanced)     & SuperBPE & Clean       & balanced     & NFC & $128{,}004$ & C \\
SuperBPE Punct t64k (bal.-half)  & SuperBPE & Punct       & balanced-half & --- & $128{,}004$ &  \\
SuperBPE/PA-BPE GPT-4o (balanced) & SuperBPE/PA-BPE & GPT-4o & balanced    & NFC & $128{,}004$ & B \\
SuperBPE/PA-BPE GPT-4o t64k       & SuperBPE/PA-BPE & GPT-4o & balanced    & NFC & $128{,}004$ & B \\
\midrule
PA-BPE GPT-4o (balanced)          & PA-BPE & GPT-4o       & balanced       & NFC & $127{,}826$ & A \\
PA-BPE GPT-4o hybrid-window (bal.)& PA-BPE & GPT-4o       & balanced       & NFC & $127{,}826$ & A \\
PA-BPE Clean (balanced)          & PA-BPE & Clean       & balanced       & NFC & $127{,}836$ &   \\
PA-BPE GPT-4o (fw2full)           & PA-BPE & GPT-4o       & fw2full        & NFC & $127{,}825$ & E \\
PA-BPE GPT-4o hybrid-win.\ (fw2full) & PA-BPE & GPT-4o    & fw2full        & NFC & $127{,}825$ & E \\
PA-BPE Clean (fw2full)           & PA-BPE & Clean       & fw2full        & NFC & $127{,}835$ & F \\
PA-BPE Clean hybrid-win.\ (fw2full) & PA-BPE & Clean    & fw2full        & NFC & $127{,}835$ & F \\
\bottomrule
\end{tabular}}
 \end{table}

\subsection{Pretokenization regexes}
\label{app:pretok}
A pretokenizer splits raw text into chunks before the subword algorithm merges
characters within chunks; it never merges across chunk boundaries, so it sets an
upper bound on which tokens can form. The 54 tokenizers use the following
pretokenizers. All of them are byte-level except the Whitespace UnigramLM tokenizer and the three SCRIPT tokenizers. The Whitespace UnigramLM tokenizer uses the Metaspace pretokenizer with byte fallback, and the SCRIPT tokenizers use the script-class encoding.

\begin{itemize}
\item \textbf{GPT-4o} (the \texttt{o200k} regex): splits text into CamelCase-aware
runs of letters with optional English contraction suffixes (\texttt{'s}, \texttt{'t},
and so on), runs of one to three digits, runs of punctuation (optionally preceded by
a single space), and whitespace. It is Unicode-property aware
(\texttt{\textbackslash p\{L\}}, \texttt{\textbackslash p\{N\}}) and is the default
for recent OpenAI tokenizers.
\item \textbf{GPT-2}: the original GPT-2 regex applied by the byte-level
pretokenizer. Similar in spirit to GPT-4o but older and ASCII-oriented; it does not
cap digit runs at three, so long numbers are grouped differently.
\item \textbf{Claude}: reverse-engineered from the Claude tokenizer. A close variant
of GPT-4o that differs in \rrev{several ways, among them the following three}: a word takes only a leading space as prefix
(not any leading non-letter, non-digit character); punctuation runs do not absorb
trailing newlines; and whitespace is split by type (spaces, tabs, newlines, and
carriage returns each grouped separately) rather than as one whitespace class. It
shares GPT-4o's CamelCase grouping and its one-to-three-digit groups. 
\item \textbf{Claude V2}: a revision of the Claude pattern from a re-probe of the
production tokenizer, used by one balanced BPE tokenizer together with an NFD
normalizer (\cref{tab:grid}). It drops the CamelCase clauses (a word is a single
run of letters and combining marks), makes the contraction suffix its own clause,
and attaches a leading space to the following letter or punctuation run only when
the preceding character is not a digit, whitespace, Han, or Hangul.
\item \textbf{Clean}: a simplified multilingual
variant of the GPT-4o pattern with a space-only word prefix that splits each digit
individually rather than in groups.
\item \textbf{Clean no-mark}: the Clean pattern with combining marks
(\texttt{\textbackslash p\{M\}}) removed from the letter classes, so a combining
mark no longer stays with its base letter and splits off into the punctuation
branch.
\item \textbf{Clean plus2 / plus3}: the Clean pattern with an attachment class on
the word branches: a leading space, Tibetan tsheg, or ASCII or curly apostrophe
attaches to the following word. Plus3 additionally attaches a word-final
apostrophe not followed by a letter. Both cap punctuation and whitespace runs at
16 characters.
\item \textbf{RightAlign}: GPT-4o modified so that multi-digit numbers split into
three-digit groups aligned from the right (``123456'' becomes ``123'', ``456''
rather than a left-aligned grouping), intended to align place value for arithmetic.
\item \textbf{Punctuation}: isolates every punctuation character as its own chunk
before the byte-level regex, so punctuation never merges with adjacent text.
\item \textbf{Whitespace}: the minimal pretokenizer. It splits only on
whitespace, with no further regex, so a chunk is a whole whitespace-delimited word;
case, digit, and punctuation boundaries inside a word are left to the subword
algorithm. The two Whitespace BPE tokenizers use this pattern. The Whitespace UnigramLM tokenizer instead uses the Metaspace pretokenizer, which splits only on spaces and not on tabs or line breaks.
\end{itemize}

\subsection{Model Architecture and Training Details}
\label{app:training}

\paragraph{Architecture.}
All models use the nanochat d24 architecture, a decoder-only transformer with 24 layers totaling 1.27B parameters total at our 128K vocabulary size. Each model is trained on 4 NVIDIA GH200 GPUs in \texttt{bfloat16}, with a context length of 2048, resulting in an effective batch size of $2^{20}=1{,}048{,}576$ tokens. 
The architecture in full: hidden dimension 1536, 12 attention heads (head
dimension 128), RoPE \citep{su2024roformer} with base frequency
$\theta = 100{,}000$, squared-ReLU activations, parameterless RMSNorm, QK-norm
with scale 1.2, and logit soft-capping at $15\cdot\tanh(\cdot/15)$; attention
uses an ``SSSL'' sliding-window pattern (three windowed layers of
$\lceil L/4\rceil$ tokens followed by one full-context layer) and value
embeddings on alternating layers; the 1.27B parameters divide into 682M
transformer matrices and 197M each for input embeddings, output projection,
and value embeddings. 

\paragraph{Optimization.}
Weight matrices use Muon \citep{jordan2025muon};
embeddings, projections, and scalars use AdamW; learning rates transfer across
widths with $\mu$P \citep{yang2022tensor} and are scaled by
$\sqrt{B/B_{\text{ref}}}=\sqrt{2}$ at the batch size $B = 2^{20}$. The learning rate warms up over 40 steps, holds
to 35\% of training, then decays linearly to 5\% of peak; Muon momentum runs
$0.85\to0.97$ over 400 steps and back to $0.90$ during warm-down; weight decay
decays cosine to zero. 
Per-group
effective settings are reported in \cref{tab:optimizer}. 

We set the token budget to nanochat's scaling-law ratio of $10.5$ tokens per parameter in the transformer matrices and the output projection. This gives $\approx 9.2$B tokens, or about $8{,}800$ steps ($8{,}794$ to $8{,}843$, depending on the vocabulary size). 
In each run, the data loader makes between about $2.0$ and $2.9$ passes over the documents of the language-model training mixture; we call one read of the mixture by the data loader a \emph{pass}, and the pass counts stated elsewhere in the paper are these loader passes. The data loader crops a document that does not fit the space that remains in a sequence. The text that a model is trained on therefore amounts to between about $1.1$ and $2.0$ times the text of the mixture (\cref{app:inversion} measures the effect of this repetition).

The smaller ranking-proxy models (nanochat d8, d12, and d16: 8, 12, and 16
layers; hidden width 512, 768, 1024; 4, 6, 8 heads; $0.22$B, $0.38$B, $0.60$B total
parameters; \cref{tab:proxy-models}) share the d24 data, tokenizers, and
optimizer recipe, with $\mu$P transferring learning rates across width, and a budget of about $3.5$B tokens. The d8 and d12 models are trained for $3{,}337$ steps, and the d16 models for $3{,}337$ to $3{,}370$ steps, depending on the vocabulary size.
 Their absolute BPB is
not comparable to the d24 models (smaller model, smaller token budget); only
the induced tokenizer ranking is used, which is invariant to a shift in BPB
level. 

\begin{table}[h]
\centering
\caption{Effective learning rates and optimizer settings per parameter group of the d24 models. $^{*}$Muon multiplies the learning rate of a matrix by the square root of its ratio of rows to columns when that ratio exceeds one. The learning rate is therefore $0.0566$ for the MLP input projection and $0.0980$ for the matrices that project the value embeddings to the attention values. The smear parameters mix the embedding of the previous token into the current position, and the backout parameter subtracts the residual stream of the middle layer before the output layer.
}
\label{tab:optimizer}
\small
\begin{tabular}{lllcl}
\toprule
\textbf{Group} & \textbf{Optimizer} & \textbf{Eff.\ LR} & \textbf{Betas} & \textbf{WD} \\
\midrule
Transformer matrices & Muon  & 0.0283{$^{*}$} & mom.\ schedule & 0.050$\to$0 \\
Input embeddings     & AdamW & 0.2121 & (0.8, 0.995) & 0.001 \\
Output projection    & AdamW & 0.0057 & (0.8, 0.96)  & 0.01 \\
Value embeddings     & AdamW & 0.1061 & (0.8, 0.995) & 0.01 \\
Residual scalars     & AdamW & 0.0071 & (0.8, 0.95)  & 0.05 \\
Skip scalars         & AdamW & 0.707  & (0.96, 0.95) & 0.0 \\
{Smear and backout parameters} & \rrev{AdamW} & \rrev{0.2} & \rrev{(0.8, 0.95)} & \rrev{0.0} \\
\bottomrule
\end{tabular}
 \end{table}

\begin{table}[h]
\centering
\caption{Smaller proxy models used for ranking transfer (\cref{tab:proxy}). All
share the d24 training data, tokenizers, optimizer recipe, and a \rrev{budget of about $3.5$B tokens, which is $3{,}337$ steps for d8 and d12 and $3{,}337$ to $3{,}370$ steps for d16}; $\mu$P transfers learning rates
across width. Parameters are totals at the 128K vocabulary.
}
\label{tab:proxy-models}
\begin{tabular}{lcccc}
\toprule
\textbf{Model} & \textbf{Layers} & \textbf{Width} & \textbf{Heads} & \textbf{Params} \\
\midrule
nanochat d8  & 8  & 512  & 4  & $0.22$B \\
nanochat d12 & 12 & 768  & 6  & $0.38$B \\
nanochat d16 & 16 & 1024 & 8  & $0.60$B \\
nanochat d24 (main) & 24 & 1536 & 12 & $1.27$B \\
\bottomrule
\end{tabular}
 \end{table}

\subsection{Model Training Data}
\label{app:data-detail}\label{sec:data}
For language-model training data, we use the same corpora as for tokenizer training, in a fixed composition unless otherwise stated, so that no comparison between two tokenizers is also a comparison between two text domains.
Matching the two keeps the tokenizer's training distribution aligned with the text it later segments.
We follow common pretraining practices, training on English and multilingual web text with smaller shares of math and coding. The mixture is 36.9\% English (FineWeb-Edu), 33.4\%
multilingual over 30 languages (FineWeb2), 16.1\% math (FineMath), and 13.6\%
code (StarCoderData), totaling 25\,GB of UTF-8 text.

Per-language data is sampled in proportion to its presence in FineWeb2; the resulting realized byte share is the resource variable of the sensitivity analysis in \cref{sec:sensitivity}.
Documents are drawn by source weight, without reference to the tokenizer, so the mixture fixes each language's share of the \emph{drawn} bytes. 
Realized byte shares of the 25\,GB mixture range from English ($36.9\%$) and Russian ($18.1\%$) to Catalan ($0.06\%$), with full break-down in \cref{tab:lang}. 
The mixture does not fix the training bytes: the BOS-bestfit packer discards a tokenizer-dependent \brev{$30.4\%$ to $43.5\%$} of the drawn text (median \brev{$35.5\%$} over the main census runs),
and the discard redistributes the per-language shares of the training bytes tokenizer-dependently rather than by one overall factor: the ratio of a language's share of the training bytes to its share of the drawn bytes deviates from one by up to \brev{$+38\%$ (Chinese, under the balanced UnigramLM tokenizer) and $-23\%$ (Hungarian, under the SuperBPE tokenizer) across five tokenizers trained on the balanced tokenizer-training data, and by $-41\%$ (Tamil) under the English-only tokenizer}.

\begin{table}[h]
\centering
\caption{Per-language data shares of the multilingual portion of the model training
mixture: GB of UTF-8 text, realized \% of the 25\,GB total, and the nominal
document-draw weight (w) the sampler used. We draw each document by first choosing a source with probability proportional to its weight and then reading that source's next document. The per-language weights are proportional to estimated character counts. The realized byte shares therefore differ from the weights where mean document length differs across sources.
Codes are those
of the training-data space; the FLORES+ evaluation pairs cmn\_Hani with
cmn\_Hans. All values from the built mixture's own metadata.}
\label{tab:lang}
\small
\adjustbox{max width=\textwidth}{%
\begin{tabular}{lllrrr@{\hspace{1.6em}}lllrrr}
\toprule
\textbf{Language} & \textbf{Code} & \textbf{Script} & \textbf{GB} & \textbf{\%} & \textbf{w} &
\textbf{Language} & \textbf{Code} & \textbf{Script} & \textbf{GB} & \textbf{\%} & \textbf{w} \\
\midrule
Russian    & \texttt{rus\_Cyrl} & Cyrillic & 4.521 & 18.1 & 0.1064 & Hungarian  & \texttt{hun\_Latn} & Latin    & 0.099 & 0.4 & 0.0051 \\
Spanish    & \texttt{spa\_Latn} & Latin    & 0.423 & 1.7 & 0.0245 & Vietnamese & \texttt{vie\_Latn} & Latin    & 0.109 & 0.4 & 0.0043 \\
German     & \texttt{deu\_Latn} & Latin    & 0.353 & 1.4 & 0.0238 & Czech      & \texttt{ces\_Latn} & Latin    & 0.077 & 0.3 & 0.0043 \\
French     & \texttt{fra\_Latn} & Latin    & 0.364 & 1.5 & 0.0206 & Greek      & \texttt{ell\_Grek} & Greek    & 0.108 & 0.4 & 0.0035 \\
Chinese    & \texttt{cmn\_Hani} & CJK      & 0.326 & 1.3 & 0.0191 & Thai       & \texttt{tha\_Thai} & Thai     & 0.087 & 0.3 & 0.0023 \\
Japanese   & \texttt{jpn\_Jpan} & CJK      & 0.204 & 0.8 & 0.0127 & Finnish    & \texttt{fin\_Latn} & Latin    & 0.034 & 0.1 & 0.0023 \\
Italian    & \texttt{ita\_Latn} & Latin    & 0.165 & 0.7 & 0.0119 & Slovak     & \texttt{slk\_Latn} & Latin    & 0.028 & 0.1 & 0.0020 \\
Portuguese & \texttt{por\_Latn} & Latin    & 0.189 & 0.8 & 0.0119 & Bulgarian  & \texttt{bul\_Cyrl} & Cyrillic & 0.052 & 0.2 & 0.0019 \\
Turkish    & \texttt{tur\_Latn} & Latin    & 0.152 & 0.6 & 0.0094 & Korean     & \texttt{kor\_Hang} & Hangul   & 0.029 & 0.1 & 0.0016 \\
Indonesian & \texttt{ind\_Latn} & Latin    & 0.236 & 0.9 & 0.0091 & Croatian   & \texttt{hrv\_Latn} & Latin    & 0.026 & 0.1 & 0.0017 \\
Polish     & \texttt{pol\_Latn} & Latin    & 0.107 & 0.4 & 0.0088 & Catalan    & \texttt{cat\_Latn} & Latin    & 0.014 & 0.1 & 0.0010 \\
Ukrainian  & \texttt{ukr\_Cyrl} & Cyrillic & 0.211 & 0.8 & 0.0068 & Hindi      & \texttt{hin\_Deva} & Devanagari & 0.032 & 0.1 & 0.0010 \\
Dutch      & \texttt{nld\_Latn} & Latin    & 0.071 & 0.3 & 0.0065 & Hebrew     & \texttt{heb\_Hebr} & Hebrew   & 0.025 & 0.1 & 0.0009 \\
Romanian   & \texttt{ron\_Latn} & Latin    & 0.097 & 0.4 & 0.0058 & Bengali    & \texttt{ben\_Beng} & Bengali  & 0.024 & 0.1 & 0.0007 \\
Arabic     & \texttt{arb\_Arab} & Arabic   & 0.168 & 0.7 & 0.0054 & Tamil      & \texttt{tam\_Taml} & Tamil    & 0.017 & 0.1 & 0.0004 \\
\bottomrule
\end{tabular}}
 \end{table}

\subsection{Evaluation and Analysis Details}\label{app:methods}
\paragraph{BPB computation.}
To compute BPB, we encode a text separately with the run's tokenizer, prepend a beginning-of-sequence token, and compute the model's loss on every token of the sentence.
We then divide by the number of UTF-8 bytes in those sentences. 
FLORES+ is not part of the model or tokenizer training data. We use the devtest split. 
Every FLORES+ sentence is shorter than the model's context of $2{,}048$ tokens under every tokenizer, so no truncation occurs and all tokenizers are scored on the same text. 
We compute Validation BPB analogously, but validation documents may be cropped when they do not fit into the remaining space in a sequence.
Because these crop locations can differ across tokenizers, we divide by the number of UTF-8 bytes in the text actually scored by each model.
The three tokenizers whose input is pre-tokenized with the encoding based on Unicode script classes have no exact per-token byte table, so no Validation BPB statistic in this paper includes them.

\begin{figure}[!htbp]
\centering
\includegraphics[width=\textwidth]{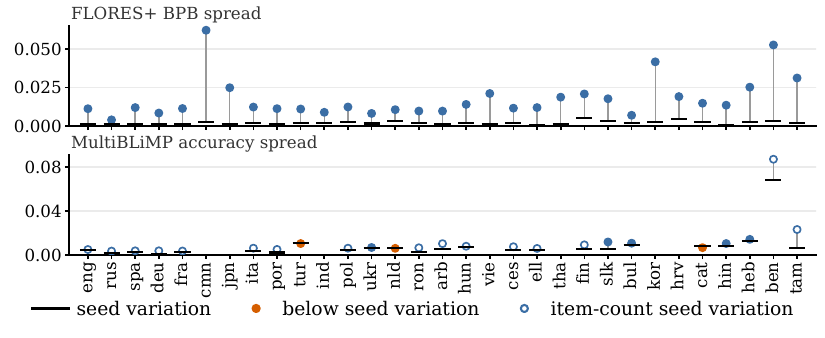}
\caption{Per-language spread of FLORES+ BPB (top) and MultiBLiMP accuracy (bottom) across the 54 tokenizers, languages \rrev{in approximately decreasing order of} share of the language-model training data (largest at left).
Each dot is the standard deviation of the outcome across the 54 tokenizers, and the horizontal tick in the same column, joined to the dot by a thin line, is that language's seed variation; orange dots are languages where the cross-tokenizer standard deviation does not exceed its seed variation (MultiBLiMP: Catalan, Dutch and Turkish). MultiBLiMP accuracy for a language is estimated from a finite number of minimal pairs, so a fixed model's measured accuracy varies with a binomial standard error of $\sqrt{(p(1-p)/n)}$ for n pairs. An open dot marks a language for which the pooled seed standard deviation does not exceed this standard error: for these languages, run-to-run variation cannot be measured more finely than the benchmark's own sampling noise. 
Languages without MultiBLiMP coverage have no dot in the bottom panel.}
\label{fig:magnitude}
\end{figure}

\paragraph{Seed variation computation.}
Retraining a model with the same tokenizer from a different random initialization changes its results slightly.
We call the size of this change the \emph{seed variation}.
For each outcome (BPB, accuracy, etc.), we estimate it as the pooled standard deviation of that outcome across seed replicates of the same tokenizer. 
The seed changes only the weight initialization; the tokenizer, training data and their order, and hyperparameters are identical across seeds.
We do this for five tokenizers: the GPT-4o, Claude, and GPT-4o-NFC balanced BPE tokenizers, the GPT-4o balanced UnigramLM tokenizer, and the GPT-4o SuperBPE tokenizer.
For each tokenizer we use only runs trained in the same period, so that differences between runs of one tokenizer come from the random seed alone.
For each outcome, we first compute its standard deviation across seed runs within each tokenizer. For per-language outcomes, we do this separately for each tokenizer--language pair. We then pool these standard deviations, weighting each by its degrees of freedom (the number of seed runs minus one). For an outcome with one value per run, the five tokenizers contribute seven degrees of freedom in total.
The seed-variation estimates used in the paper differ only in the outcome and in what is pooled:
\begin{itemize}[leftmargin=15pt,noitemsep]
    \item \emph{Trained languages}: FLORES+ BPB for each of the 31 trained languages, pooled over the $155$ tokenizer--language pairs (5 tokenizers $\times$ 31 languages).

    \item \emph{Single language}: that language's FLORES+ BPB, pooled over the five tokenizers; these values are shown in \cref{fig:magnitude}.

    \item \emph{31-language mean}: for each run, we first average FLORES+ BPB over the 31 trained languages, then compute the standard deviation across seeds within each tokenizer and pool over the five tokenizers.

    \item \emph{Validation BPB}: computed and pooled in the same way over the five tokenizers.
\end{itemize}

The main analyses use the 31 trained languages; the fixed 214-language FLORES+ set of \cref{sec:setup} adds 183 languages that are mostly unseen during training; while we don't explicitly include these languages in our corpora, misidentified or code-switched text from these languages may appear depending on the language filtering.  
On these languages the run-to-run noise is larger: the per-language seed variation
over all 214 languages is \rrev{$0.0146$},
about \rrev{six} times the seed variation of $\rrev{0.0023}$ over the languages in model-training data.

For the 31 languages of the main analysis, the estimated seed variation is $\rrev{0.0023}$ for a single language's FLORES+ BPB,
$0.0012$ for the 31-language FLORES+ mean,
and $0.00041$ for validation BPB. 

\paragraph{Standard error computation.}
The 54 tokenizers are not independent observations: they fall into $G=42$ families, and tokenizers from the same family are likely to behave similarly (\cref{sec:setup}). The model-based standard errors and $p$-values from the mixed-effects fits do not account for this dependence and can therefore be too small.
We instead use clustered standard errors, estimated from an ordinary-least-squares fit of the same within-language specification with a fixed effect per language and standard errors clustered at the tokenizer-family level ($G=42$; \cref{sec:setup}).
For the analyses looking at the relationships between intrinsic metrics and downstream outcomes, the OLS coefficients \rrev{agree with} the mixed-effects coefficients \rrev{to within one unit of the last digit} printed in \cref{tab:within}.
Clustering inflates the standard error of each metric by a factor between $1.0$ and \rrev{$3.2$} relative to the model-based standard error, with the smallest inflation for the UTF-8 boundary-crossing rate; we therefore rely on the clustered $p$-values throughout.
The tier, log-weight-interaction, and typology fits of \cref{sec:metrics} use the same construction: an OLS fit with language fixed effects and the metric-by-attribute interaction term, with standard errors clustered by family and BH adjustment within each block of tests. The attribute main effects are absorbed by the language fixed effects.

\section{Robustness of Design Choices}\label{app:rq1-2} 
\label{app:robustness-design}

In this section, we report additional robustness checks for our main conclusions. Specifically, we examine (i) whether the relationships of a language's tokenizer sensitivity and exclusion penalty with its share of the language-model training data hold under alternative ways of measuring them; (ii) whether our results depend on how the outcome is normalized across languages; (iii) whether resource level can be separated from writing system; (iv) whether the exclusion penalty is confounded by the amount of text a model sees; and (v) whether tokenizer rankings hold beyond repeated data and a single token budget.

\paragraph{Robustness of the BPB spread--model-training data share relationship.}
The results in \cref{sec:sensitivity} suggests that languages with smaller shares of the language-model training data have larger BPB spreads across tokenizers. We test if this result is inflated by how we count tokenizers or by run-to-run seed variation.
We first observe that counting near-duplicate tokenizers separately does not create this relationship: when we compute each language's BPB spread across one mean per tokenizer family, using 42 families instead of 54 tokenizers, its Spearman correlation with log language-model training share is $\rrev{-0.54}$, vs. $\rrev{-0.52}$ at tokenizer level.

The per-language seed variation also falls as language-model training share rises: the pooled seed standard deviation of a language's FLORES+ BPB correlates with its log training share at Spearman $\rho = \rrev{-0.608}$ ($p = \rrev{0.0003}$). Because this relationship has the same direction as the BPB-spread relationship, some of the association between BPB spread and training share may reflect larger run-to-run variation for lower-resource languages.
Finally, for each language, we take the largest absolute correlation between FLORES+ BPB and any of the nine intrinsic metrics across tokenizers. This value is negatively and significantly correlated with the language's BPB spread (Spearman $\rho = -0.42$): in plain terms, the languages for which tokenizer choice changes BPB the most are not those for which any single intrinsic metric has the strongest relationship with BPB.

\paragraph{Robustness to outcome normalization.}
Bits per byte (BPB) is not a language-neutral unit~\citep{yang2026applesapplescomparablecrosslingual} as it divides the cross entropy loss by the number of UTF-8 bytes in the text the loss is computed over. While FLORES+ is a parallel corpus (i.e. each language subset contains the same content), the number of bytes in the documents can vary greatly across languages. 
It depends on the number of characters per sentence and on the number of UTF-8 bytes per character of the language's script. 
For instance, the Chinese sentences have $124$ bytes on average, their German translations $159$, and their Bengali translations $356$, which demonstrates that per-byte information content varies greatly across languages.

For this reason, we also check if dividing by the number of sentences in each document changes the conclusions drawn based on BPB computations.
Because the number of bytes per sentence is constant across tokenizers within a language, every within-language comparison is identical under the two outcomes: the tokenizer ranking, the ratio of spread to seed variation, and the sign and significance of each within-language metric coefficient.
Cross-language comparisons can differ, but the relationship between tokenizer sensitivity and language-model training share is robust to the choice of normalization. The Spearman correlation between a language's spread across tokenizers and its log share of the language-model training data is $\rrev{-0.52}$ using the standard deviation of BPB, $\rrev{-0.54}$ using the standard deviation of bits per sentence, and $\rrev{-0.38}$ BPB coefficient of variation (when spread is divided by the language's mean BPB); over the 28 languages written with word boundaries, the corresponding correlations are $\rrev{-0.69}$, $\rrev{-0.69}$, and $\rrev{-0.51}$. All are significant.
Under bits per sentence, the exclusion penalty's correlation with the log LM share is $-0.26$ and not significant, as it is under BPB (\cref{sec:rq2}). 

The identity of the \emph{most} sensitive language does depend on the normalization: Chinese has the largest BPB spread (followed by Bengali and Korean,~\cref{fig:spread-coverage}), but after normalizing BPB variance by the language's mean BPB or when measuring variance in bits per \emph{sentence} rather than bits per \emph{byte}, and Chinese is third, after Bengali and Tamil---the two lowest resource languages. 

Similarly, the predictor, a language's share of the language-model training data, is also a byte count; where equal byte counts do not correspond to equal amounts of content across languages \citep{arnett2024bytePremium}. As a second measure of a language's share, we divide its model-training bytes by the mean number of bytes of its FLORES+ sentences, which approximates the amount of its training text in FLORES+ sentences. Across the 31 languages, the Spearman correlation between the exclusion penalty and the logarithm of this measure is $-0.36$ and significant, vs. $-0.33$ and not significant for the byte share (in \cref{fig:spread-coverage}, right). 

\clara{clarify MultiBLiMP}
\clara{check what's meant by "seed variation is set by the MultiBLiMP item count" }

\paragraph{Separating resource level from writing system.}
The relationship between a language's tokenizer sensitivity and model-training share could be entangled with the underlying writing system. 
We therefore test if resource level explains a language's tokenizer sensitivity beyond its script. 
This analysis uses one observation per language, not the  $1{,}674$ tokenizer--language pairs: the pairs share the same 31 languages, so a standard error computed as if the pairs were independent would likely be an underestimate, giving false (overconfident) confidence intervals.

For the across-tokenizer BPB spread of \cref{fig:spread-coverage} (where we include one value per language), a non-Latin-script indicator alone explains more of the variance ($R^2 = 0.17$) than logarithm of the model-training data share alone ($R^2 = 0.08$), but neither predictor is stable across specifications: an added indicator for the three languages written without word boundaries is significant, and its addition makes the non-Latin indicator not, and refitting on the 28 languages with word boundaries makes the log share significant ($\rrev{-0.69}$ vs. $\rrev{-0.52}$ with all languages included).
We report this as a descriptive variance partition, not a causal decomposition: the predictors are entangled in the set of languages we explored.
For a different entangled axes, the tokenizer-training share, the resource-inversion experiment of \cref{sec:design} breaks the entanglement by construction: under the inverted-allocation tokenizer, the measured correlation of a language's log share of the tokenizer-training data with its log nominal sampling weight in the language-model training data is $-0.961$ vs. $\rrev{+0.199}$ and $\rrev{+0.133}$ under the permuted allocations. 
Given our limited pool of languages, the writing system is not a variable whose effect we can cleanly isolate\clara{come back to this}.

\paragraph{Exclusion penalty confounders.} 
The exclusion penalty result in \cref{res:rq2} cannot be attributed cleanly as an effect of tokenizer coverage: excluding a language from the tokenizer-training data also causes the model to see less training bytes from that language in most cases.
This occurs because the excluding tokenizers typically segment the language into more tokens, so a fixed training-token budget covers less training bytes.
For each of the 30 languages excluded by at least one coverage tokenizer, we compute the ratio of training bytes under the excluding tokenizers to those under the including tokenizers.
The ratio is below one for all 30 languages, with a median of $0.82$, and languages with smaller ratios have larger exclusion penalties (Spearman $\rho=-0.47$, significant).
Restricted to the 15 languages excluded by at least two coverage tokenizers, the penalty--share correlation of \cref{sec:rq2} is $-0.63$ and significant.

\paragraph{No large reordering of tokenizers appears without repeated data or at twice the training budget.}
Tokenizer rankings across the study come from language model training runs that make two to three passes over the language-model training mixture at a fixed token budget.
We test the impact of both design choices on our conclusions.
We train 11 additional models with 10 of the tokenizers in the study: one model for each of nine tokenizers from the main set, and two seeds of the inverted-allocation tokenizer.
First, we trained these 11 models on a single-pass corpus at the same token budget.
The Spearman correlation between their 31-language FLORES+ means and those of the same tokenizers' trained with our typical repeated-data setup is $\rrev{+0.93}$ ($\rrev{+0.90}$ over ten tokenizers when the two inverted-allocation seeds are averaged first).
\dealwithlater{\brev{Of the 55 pairs of runs, 7 change order, and in 4 of the 7 the repeated-data difference is larger than twice the seed variation of a two-run difference.}
}
Second, we continued to train these single-pass run for an additional 10B tokens of new text, to about twice the token budget of our typical setup.
The Spearman correlation between the eleven means before and after the continuation is $\rrev{+0.94}$.\clara{come back to this} 
In both comparisons, the BPB rank agreement is lower for single languages than for the means across languages: the lowest per-language correlations are $-0.11$ (Chinese) and $\rrev{+0.23}$ (Hindi) between single-pass and repeated data, and $\rrev{+0.64}$ (Finnish) and $\rrev{+0.83}$ (Slovak) across the continuation.
With eleven runs, a large reordering would be visible in these comparisons, and a small reordering cannot be distinguished from noise.

\section{Controlled Comparisons}
\label{app:pretok-ablation}

\begin{table}[!h]
\centering
\caption{Controlled comparisons: 25 tokenizer pairs that differ in one configuration choice.
$\Delta$FLORES: variant minus baseline in the 31-language FLORES+ mean.
$\Delta$Val: variant minus baseline in validation BPB.
The ratio column after each difference is the difference's magnitude divided by the unrounded pooled seed standard deviation of that outcome across seed replicates (\cref{app:methods}).
\textbf{Bold}: the difference is larger than run-to-run variation (\cref{sec:setup:analysis}), i.e., the ratio is at least $2\sqrt{2} \approx 2.83$; the factor $\sqrt{2}$ converts the standard deviation of one run into the standard deviation of a difference between two runs. 
$^{a}$full-byte-alphabet rebuild; $^{b}$tokenizer with the byte-drop defect (\cref{app:grid})
\dealwithlater{\brev{; $^{c}$the direction reverses when both runs are matched on training bytes rather than on training tokens (\rrev{\cref{app:exclusion-bytematch}})}.}
}
\label{tab:select}
\small
\begin{tabular}{llrrrr}
\toprule
& \textbf{Contrast (baseline $\to$ variant)} & $\Delta$\textbf{FLORES} & ratio & $\Delta$\textbf{Val} & ratio \\
\midrule
\multirow{6}{*}{\rotatebox{90}{Data}}
& Balanced $\to$ English-only (GPT-4o)$^{a}$   & \rrev{$\mathbf{+0.0386}$} & \rrev{$32.69$} & \brev{$\mathbf{+0.0332}$} & \brev{$81.73$} \\
& Balanced $\to$ English-only (Claude)$^{b}$   & \rrev{$\mathbf{+0.0307}$} & \rrev{$26.06$} & \brev{$\mathbf{+0.0248}$} & \brev{$61.00$} \\
& Balanced $\to$ English-only (Punct)$^{b}$    & \rrev{$\mathbf{+0.0238}$} & \rrev{$20.18$} & \brev{$\mathbf{+0.0254}$} & \brev{$62.41$} \\
& Balanced $\to$ Code-heavy (GPT-4o)           & \rrev{$\mathbf{+0.0189}$} & \rrev{$16.06$} & \brev{$\mathbf{+0.0153}$} & \brev{$37.56$} \\
& Balanced $\to$ High-resource (GPT-4o)        & \rrev{$\mathbf{+0.0202}$} & \rrev{$17.09$} & \brev{$+0.00025$} & \brev{$0.62$} \\
& Balanced $\to$ High+mid (GPT-4o)             & \rrev{$\mathbf{+0.0121}$} & \rrev{$10.29$} & \brev{$+0.00059$} & \brev{$1.45$} \\
\midrule
\multirow{6}{*}{\rotatebox{90}{Algorithm}}
& BPE $\to$ UnigramLM (GPT-4o)                 & $\mathbf{+0.0323}$ & \rrev{$27.41$} & \brev{$\mathbf{+0.0199}$} & \brev{$48.92$} \\
& BPE $\to$ UnigramLM (Claude)                 & \rrev{$\mathbf{+0.0285}$} & \rrev{$24.19$} & \brev{$\mathbf{+0.0198}$} & \brev{$48.81$} \\
& BPE $\to$ UnigramLM (RightAlign)             & \rrev{$\mathbf{+0.0293}$} & \rrev{$24.86$} & \brev{$\mathbf{+0.0192}$} & \brev{$47.32$} \\
& BPE $\to$ SuperBPE (GPT-4o)                  & \rrev{$\mathbf{+0.0130}$} & \rrev{$11.02$} & $\mathbf{+0.0105}$ & \brev{$25.94$} \\
& PA-BPE $\to$ SuperBPE+PA-BPE (GPT-4o)%
& \rrev{$-0.0025$} & \rrev{$2.11$} & \brev{$\mathbf{+0.0035}$} & \brev{$8.51$} \\
& BPE $\to$ PA-BPE (Clean, NFC)                & \rrev{$\mathbf{+0.0200}$} & \rrev{$16.95$} & \brev{$\mathbf{+0.0086}$} & \brev{$21.11$} \\
\midrule
\multirow{6}{*}{\rotatebox{90}{Pretokenizer}}
& GPT-4o $\to$ Claude                          & \brev{$+0.0014$} & \rrev{$1.18$} & $\mathbf{+0.0024}$ & \brev{$5.89$} \\
& GPT-4o $\to$ GPT-2                           & \rrev{\brev{$-0.0018$}} & \rrev{$1.55$} & \brev{$-0.00033$} & \brev{$0.81$} \\
& GPT-4o $\to$ Punctuation                     & \rrev{$\mathbf{+0.0048}$} & \rrev{$4.07$} & $\mathbf{+0.0054}$ & \brev{$13.27$} \\
& GPT-4o $\to$ RightAlign                      & \rrev{\brev{$+0.0030$}} & \rrev{$2.52$} & \brev{$+0.00066$} & \brev{$1.62$} \\
& GPT-4o $\to$ Whitespace                      & \rrev{$\mathbf{+0.0049}$} & \rrev{$4.19$} & \brev{$\mathbf{+0.0078}$} & \brev{$19.23$} \\
& GPT-4o $\to$ Clean (PA-BPE)                   & \rrev{$\mathbf{-0.0070}$} & \rrev{$5.95$} & \brev{$\mathbf{-0.0019}$} & \brev{$4.75$} \\
\midrule
\multirow{3}{*}{\rotatebox{90}{NFC}}
& GPT-4o $\to$ GPT-4o NFC                      & \rrev{\brev{$-0.0019$}} & \rrev{$1.65$} & $+0.00022$ & $0.55$ \\
& Claude $\to$ Claude NFC                      & \rrev{\brev{$-0.0036$}} & \rrev{$3.08$} & \brev{$-0.00033$} & \brev{$0.81$} \\
& RightAlign $\to$ RightAlign NFC              & \rrev{\brev{$-0.0038$}} & \rrev{$3.22$} & \brev{$+0.00002$} & \brev{$0.05$} \\
\midrule
\multirow{4}{*}{\rotatebox{90}{Presets}}
& UnigramLM default $\to$ tuned (GPT-4o)       & \rrev{$-0.0004$} & \rrev{$0.33$} & \brev{$-0.00002$} & \brev{$0.06$} \\
& PA-BPE base $\to$ hybrid-window (GPT-4o)      & \rrev{$\mathbf{-0.0084}$} & \rrev{$7.11$} & \brev{$\mathbf{-0.0065}$} & \brev{$15.88$} \\
& SuperBPE+PA-BPE t90k $\to$ t64k              & \rrev{$-0.0004$} & \rrev{$0.37$} & \brev{$-0.00008$} & \brev{$0.20$} \\
& SuperBPE Clean C2 $\to$ C3                   & \rrev{$\mathbf{+0.0044}$} & \rrev{$3.75$} & \brev{$\mathbf{+0.0012}$} & \brev{$2.88$} \\
\bottomrule
\end{tabular}
 \end{table}

\cref{tab:select} presents the results used in \cref{sec:rq3}. Each row compares two tokenizer--model runs whose tokenizers differ in one configuration choice while the language-model architecture, optimization, training mixture, and training-token budget are held fixed.
Thus, the comparisons isolate a tokenizer configuration change under our standard training setup. 
Notably, they do not hold fixed the amount of training bytes seen during language-model training: changing the tokenizer can change compression, so models trained for the same number of tokens may see different numbers of bytes. 

The largest change of the 25 comes from restricting the tokenizer-training data to English: it raises both the 31-language FLORES+ mean BPB and validation BPB significantly across three different pretokenization regexes. 
The milder restrictions, to the high-resource languages or to the high-and-mid-resource languages, similarly negatively affect performance but to a lesser extent (and not significantly when measured on validation BPB.
Of the five pretokenizer-regex changes measured against the GPT-4o baseline, the two largest are the Whitespace pretokenizer (+0.0049 on the 31-language FLORES+ mean) and isolating every punctuation character (+0.0048). 
NFC normalization lowers the 31-language FLORES+ mean in all three pairs, in two of them (Claude and RightAlign) by more than run-to-run variation. 
Choice of algorithm also significantly impacts BPB in most cases. 
BPB consistently increases when switching to the UnigramLM tokenization algorithm. 
The HuggingFace UnigramLM trainer used here is documented to differ from, and perform worse than, the original SentencePiece implementation, though. Part of the UnigramLM difference may be the implementation rather than the algorithm, and so we do not read strongly into these results.
\dealwithlater{Part of each algorithm difference is a difference in training bytes rather than in the algorithm: at the fixed token budget the UnigramLM variant trains on fewer bytes than its balanced BPE baseline and the SuperBPE variant on more, and retraining each to that baseline's byte count halves the UnigramLM difference, to $+0.0182$, and grows the SuperBPE difference, to $+0.0148$; both remain far above the seed variation of the 31-language FLORES+ mean (\cref{app:pretok-ablation}).\brev{ When we hold training bytes fixed instead of training tokens, the direction of one comparison of \cref{tab:select} reverses: adding the SuperBPE stage to parity-aware BPE changes the 31-language FLORES+ mean by $-0.0049$ at the fixed token budget and by $+0.0148$ at matched bytes, the same direction and size as for plain BPE (\cref{app:pretok-ablation}).}\footnote{The HuggingFace UnigramLM trainer used here is documented to differ from, and perform worse than, the original SentencePiece implementation, so part of the UnigramLM difference may be the implementation rather than the algorithm.}}
\paragraph{A byte-alphabet defect lowers measured BPB slightly and does not produce any of the reported BPB increases.}
When BPE training is run without an explicit initial byte alphabet, every byte value absent from the fitting corpus will be missing from the vocabulary, and text containing such a byte cannot be fully encoded. This happened for several of the tokenizers in this study, in particular, those trained on only English data. For these defective tokenizers, 32 of the 128 tested byte values are missing: every C0 control character except tab and newline, the left brace, and the delete character. One pair isolates the defect's impact: the English-data GPT-4o BPE tokenizer of \cref{tab:grid} is a defect-free rebuild of a defective original, identical in algorithm, pretokenization regex, and tokenizer-training data. The rebuild's 31-language FLORES+ mean is higher than the defective original's by $0.009$ and its validation BPB by $0.003$; per language, the rebuild's FLORES+ BPB is higher for 30 of the 31 languages, and for the one language where it is lower, the difference is within seed variation. The defect therefore ostensibly \emph{lowers} measured BPB.\clara{Need to comment that this might be measurement confound} The exclusion penalties of \cref{sec:sensitivity} use the rebuild and involve no defective tokenizer, and for the two defective BPE tokenizers retained in the 54-tokenizer panel the defect lowers their measured BPB rather than raising it.

\section{Equal Weight and Resource-inversion Experiments}
\label{app:inversion}
In \cref{sec:rq2}, we find that a language's share of language-model training data negatively correlates with the spread of its FLORES+ BPB across tokenizers, i.e., lower-resource languages are more sensitive to tokenizer choice. 
Most tokenizers in the panel (\cref{tab:grid}) are trained on the \textit{balanced} corpus, where a language's share of the tokenizer-training data is proportional to its share of the language-model training data, so resource-level for language-model and tokenizer training data are entangled. 
The observed correlation could potentially be an artifact of this. 
We isolate the impact of tokenizer-training share by reallocating it independently of the language-model training share: we invert the natural allocation, permute it in two ways, exclude languages from it, and weight all languages equally (\cref{app:sec:inversion:setup}). \Cref{app:sec:inversion:reallocations} reports the correlations between the tokenizer-training data share and changes to FLORES+ BPB under these controlled experiments; \cref{app:sec:inversion:further} rules out three alternative explanations for the trends that we observe; and \cref{app:sec:inversion:compared} compares the results with the exclusion result of \cref{sec:sensitivity}.

\subsection{Set-up}
\label{app:sec:inversion:setup}
\paragraph{Equal weighting variants}
For the equal allocation ablations, we train five additional tokenizers where the tokenizer-training weights are equal across 30 non-English languages, and the rate of the English data is held fixed at 35\%. 

\paragraph{Reference and variant runs for inversion and permutation experiments.}
We fix the balanced GPT-4o BPE tokenizer and its corresponding model as the reference tokenizer--model pair for the inversion and permutation experiments\dealwithlater{. The reference model is a separate run with this tokenizer, in addition to the run of the balanced GPT-4o BPE tokenizer among the 54 tokenizers} (\cref{sec:design}).
We extend our analysis to four additional settings that differ from the reference tokenizer only in the per-language weights of the tokenizer-training data; 
\dealwithlater{\rrev{and in the limit on the number of files read from the English web text, math, Python, and JavaScript sources. The} }pretokenizer, normalizer, algorithm, vocabulary size, and special tokens are held constant.
We label the four variants by the released artifacts: L2 is trained on inverted-allocation tokenizers, P1 and P2 are the two trained on permuted allocation tokenizers, and E1 is the variant trained without English web text, each elaborated below.
We train the L2 model under two random seeds and average the two runswherever both are available.

This change affects the exact realized shares (measured in UTF-8 bytes) of the language-model training data. 
In general, a tokenizer trained on less of a language segments that language into more tokens, so at a fixed number of steps its model is trained on fewer bytes of text.
To control for this in these experiments, we therefore set each variant's number of steps so that its model is trained on the same number of bytes as the reference model: between $9{,}087$ and $9{,}980$ steps vs. $8{,}800$ for the reference model.

\begin{table}[!htbp]
\centering
\caption{Equal vs. proportional tokenizer-data weighting across 30 non-English languages (English fixed at 35\%).
$\Delta$ mean: the 31-language FLORES+ mean of the model with equal weights minus that of the model with proportional weights.
$\Delta$ tail: the same difference in FLORES+ BPB, averaged over the 15 languages with the smallest proportional weights.
Spearman $\rho$: correlation between a language's change in FLORES+ BPB and its proportional weight.
Bold and a star mark correlations with $p < 0.05$.
\rrev{Last row: a control that changes only the limit on the number of files.}}
\label{tab:equalweight}
\small
\begin{tabular}{lccc}
\toprule
\textbf{Variant (vs.\ balanced baseline)} & $\Delta$ \textbf{mean} & $\Delta$ \textbf{tail} & \textbf{Spearman} $\rho$ \\
\midrule
Equal weights (GPT-4o)             & \rrev{$+0.0107$} & \rrev{$+0.0177$} & $\mathbf{-0.61}^{*}$ \\
Equal weights (Claude)             & \rrev{$+0.0064$} & \rrev{$+0.0162$} & \rrev{$\mathbf{-0.52}^{*}$} \\
Equal weights (Punct)              & \rrev{$+0.0015$} & \rrev{$+0.0031$} & \rrev{$-0.07$} \\
Equal weights (GPT-4o, NFC)        & \rrev{$+0.0054$} & \rrev{$+0.0099$} & \rrev{$-0.34$} \\
Equal weights, no repeat sampling  & \rrev{$+0.0060$} & \rrev{$+0.0120$} & \rrev{$\mathbf{-0.44}^{*}$} \\
\midrule
File cap only (control)            & \rrev{$-0.0035$} & \rrev{$-0.0039$} & \rrev{$+0.17$} \\
\bottomrule
\end{tabular}
 \end{table}

\subsection{Reallocation Effects}
\label{app:sec:inversion:reallocations}
\paragraph{Equal-weighting.} 
We compare each equal-weight tokenizer with the tokenizer trained on the balanced data that matches it in every other configuration choice in
\Cref{tab:equalweight}.
In every pair the 31-language FLORES+ mean and the mean over the 15 languages with the smallest proportional weights are both \textit{higher} under the equal weights, and in four of the five pairs the rise in the 31-language mean is larger than run-to-run variation.

\paragraph{Coverage tokenizers.}
In \Cref{sec:sensitivity}, we examine each language's exclusion penalty on a coverage tokenizer trained without it. \Cref{tab:exclusion-general-difference} reports the difference in FLORES+ BPB across subsets of the 31 languages.

\begin{table}[t]
\centering
\caption{For the three groups of languages that the same coverage tokenizers exclude: the difference between the coverage tokenizers that is not specific to a language. 
Difference in 31-language mean: the average of the 31-language FLORES+ means of the coverage tokenizers that exclude the group's languages, minus the average of the 31-language FLORES+ means of the coverage tokenizers that include them.
Mean penalty: the mean of the exclusion penalties (\cref{sec:sensitivity}) of the group's languages.
Ratio: the difference in 31-language mean divided by the mean penalty.
English is not included in the analysis below. \dealwithlater{in any group: its penalty of $\rrev{+0.0012}$ compares the no-English variant with the four coverage tokenizers, and the 31-language FLORES+ mean of the no-English variant is lower than the average of the 31-language FLORES+ means of the four coverage tokenizers by $\rrev{0.015}$.}}
\label{tab:exclusion-general-difference}
\small
\begin{tabular}{p{0.34\linewidth}rrrr}
\toprule
\textbf{\# of coverage tokenizers that exclude the language} &\shortstack[r]{\textbf{\# of languages }\\\textbf{in group}}  & \shortstack[r]{\textbf{Difference in}\\\textbf{31-language mean}} & \shortstack[r]{\textbf{Mean}\\\textbf{penalty}} & \textbf{Ratio} \\
\midrule
One (English-only) & 5 & \rrev{$0.0278$} & \rrev{$0.0671$} & $0.41$ \\
Two (English-only, high-resource) & 15 & \rrev{$0.0233$} & $0.0213$ & \rrev{$1.09$} \\
Three (English-only, high-resource, high-and-mid-resource) & 10 & \rrev{$0.0236$} & \rrev{$0.0330$} & \rrev{$0.72$} \\
\bottomrule
\end{tabular}

 \end{table}

\paragraph{Inverted and permuted allocations}
\Cref{tab:inversion-arms} reports, for each of the four variants, the Spearman correlation between a language's change in FLORES+ BPB against the reference run and the language's realized log share of the language-model training data.
The inverted-allocation tokenizer is the only variant whose correlation is significant, that is, with $p < 0.05$.
The difference between the inverted-allocation tokenizer's correlation and each permuted-allocation tokenizer's correlation is tested by a bootstrap, which re-draws the 31 languages with replacement and recomputes both correlations on each draw.
The difference with respect to permuted allocation 1 is $-0.67$ (significant) and with respect to permuted allocation 2 is $-0.30$ (not significant).

\begin{table}[h]
\centering
\caption{Spearman correlation, for each reallocated tokenizer, between a language's change in FLORES+ BPB against the reference run and the language's log share of the language-model training data (realized share), over the 31 trained languages. A star marks $p < 0.05$.}
\label{tab:inversion-arms}
\small
\begin{tabular}{llr}
\toprule
Label & Variant & Spearman $\rho$ \\
\midrule
L2 & inverted-allocation tokenizer (two seeds averaged) & $\rrev{-0.59}^{*}$ \\
P1 & permuted allocation 1 & $\rrev{+0.08}$ \\
P2 & permuted allocation 2 & $\rrev{-0.28}$ \\
E1 & no English web text & \rrev{$-0.13$} \\
\bottomrule
\end{tabular}
\end{table}

\subsection{Testing for further confounders}
\label{app:sec:inversion:further}
Before attributing Tamil's rise in FLORES+ BPB to the vocabulary itself, we test three alternative explanations: that the inverted-allocation tokenizer simply trains the model on fewer Tamil tokens, that our repeated-data training regime amplifies the effect, and that the rise is concentrated in vocabulary entries the training data contain too rarely to learn well.

\paragraph{Token matching.}
When a tokenizer segments a language into fewer tokens, the model of that tokenizer is trained on fewer tokens of that language. Under the inverted-allocation tokenizer, the model is trained on fewer Tamil tokens, even though it ends up being trained on more underlying Tamil text. Part of the increase we observe could be due to this phenomenon rather than an effect of the vocabulary itself. 

To test this, we train two additional models on a reweighted language-model training mixture, one with the reference tokenizer (BPE GPT-4o tokenizer trained on balanced mixture) and one with the inverted-allocation tokenizer. On this reweighted mixture, a model with the inverted-allocation tokenizer is trained on as many tokens of each language as the reference model is on the original mixture. 
Matching token counts this way means the model on the reweighted mixture is trained on a \emph{different amount of training bytes}, compared to the set-up described above.
\Cref{tab:tokenmatch-main} reports the four combinations of tokenizer (reference or inverted) and language-model training mixture (original, or reweighted so that the model with the inverted-allocation tokenizer is trained on as many tokens of each language as the reference model is on the original mixture; \cref{sec:design}). 

On the reweighted mixture, the model with the inverted-allocation tokenizer is $+0.181$ above the reference model, and $+0.233$ above the model with the reference tokenizer trained on the same reweighted mixture ($0.181 - (-0.052)$); this between-tokenizer difference at matched Tamil token counts is close to the $+0.267$ difference at matched steps on the original mixture. 
This suggests that \rrev{with repeated data,} the smaller number of Tamil training tokens is not the main cause of Tamil's higher FLORES+ BPB. 

\begin{table}[h]
\centering
\caption{Tamil's FLORES+ BPB minus that of the reference model on the original mixture, for the four combinations of tokenizer and language-model training mixture. Reweighted mixture: the mixture on which the model with the inverted-allocation tokenizer is trained on as many tokens of each language as the reference model is on the original mixture.}
\label{tab:tokenmatch-main}
\small
\begin{tabular}{lcc}
\toprule
& Original mixture & Reweighted mixture \\
\midrule
Reference tokenizer & $0$ & $\rrev{-0.052}$ \\
Inverted-allocation tokenizer & $\rrev{+0.267}$ & $+0.181$ \\
\bottomrule
\end{tabular}
\end{table}

\paragraph{Single-pass replication.}
Every language-model training run of the 54 tokenizers and every additional run described so far in this section makes between two and three passes over the language-model training mixture (\cref{sec:setup}).
To test whether repeating training data during language-model training matters, we train three additional models on a larger mixture of the same composition in which no document is processed twice: one run of the reference tokenizer and two seeds of the inverted-allocation tokenizer, with final checkpoint scores for the two averaged.
We contruct this single-pass mixture so that the data seen in the beginning matches the original run's first pass, document for document; the two mixtures differ only after the point at which the original run begins its second pass.
Before that point, a single-pass run and its repeated-data counterpart have the same tokenizer, data, and seed.
As a check that the two runs match until that point, we compare\rrev{, at steps $1{,}000$ and $2{,}500$, the per-language change in FLORES+ BPB under the inverted-allocation tokenizer on single-pass data with the same change on repeated data (seed 42)}\rrev{. Of the 62 language-step entries, 4} differ by more than $0.02$, and the largest difference is $0.0221$ (Turkish at step $1{,}000$).
We attribute these differences to evaluation at single checkpoints.

\Cref{tab:repetition} reports the per-language change in FLORES+ BPB under the inverted-allocation tokenizer against the reference run, on repeated data and on single-pass data.
Tamil's change is $\rrev{+0.1386}$ on single-pass data against $\rrev{+0.2674}$ on repeated data.
The pre-registered criterion for a material effect of repetition was that the single-pass change falls below $0.7$ of the repeated-data change; the criterion is met.
\rrev{For the two other languages of \cref{tab:repetition}, the change on single-pass data is within $0.01$ of zero.}

The Spearman correlation between a language's change in FLORES+ BPB and its log share of the language-model training data is $\rrev{-0.59}$ on repeated data and $\rrev{+0.34}$ on single-pass data.
\rrev{The repeated-data correlation is significant; the single-pass correlation is not}.
Computed separately for the two single-pass seeds, the correlation is $\rrev{+0.33}$ and $\rrev{+0.30}$, \rrev{where neither is significant}.
The sign of the single-pass correlation does not depend on any one language: leaving out each of the 31 languages in turn and recomputing the correlation ranges from $\rrev{+0.29}$ to $\rrev{+0.47}$.
\rrev{Using all 31 languages, the correlation is not significant, and it stays not significant when 26 of the 31 languages are held out in turn.}
\rrev{Chinese has the largest single-pass rise ($+0.14$); leaving it out lowers the correlation to $+0.29$ with $p = 0.12$.}
\dealwithlater{\rrev{We evaluated intermediate checkpoints of the seed-42 single-pass run only. In that run, the} Chinese value at the single-pass endpoint comes from a checkpoint in the middle of a within-run oscillation.}

For each language, we compute the change on repeated data minus the change on single-pass data, to quantify how much repetition hurts performance in that language. 
We find that this difference correlates with the language's log share of the language-model training data at Spearman $\rrev{-0.89}$ (significant): \emph{repetition hurts most for \rrev{Bengali and Tamil, the two} languages whose fertility ratio fell furthest under the inverted-allocation tokenizer}.
Under the reference tokenizer, the single-pass run's 31-language FLORES+ mean is lower than the repeated-data run's by $0.0164$, so \emph{single-pass data lowers BPB in general} independent of the change in allocation. The per-language difference above removes this general shift, isolating the language-specific effect of repetition.
The results of \cref{sec:design} are therefore results for repeated-data training, which every run of the 54 tokenizers shares.
On single-pass data, about half of Tamil's rise remains, and the rise is not larger for the languages with a smaller share of the language-model training data.

\begin{table}[h]
\centering
\caption{Per-language change in FLORES+ BPB from the reference run under the inverted-allocation tokenizer, on the repeated training data of the 54 tokenizers' runs and on single-pass data: the two languages whose fertility ratio fell furthest, two languages whose fertility ratio changed less, and the mean over the 31 training languages. In both settings the inverted-allocation value is the average of two seeds, compared against one reference run; the largest difference between the two seeds in any language's change is $\rrev{0.017}$ on repeated data and $\rrev{0.019}$ on single-pass data.}
\label{tab:repetition}
\small
\begin{tabular}{lcc}
\toprule
& \textbf{Repeated} & \textbf{Fresh (single epoch)} \\
\midrule
Tamil    & \rrev{$+0.2674$} & \brev{\rrev{$+0.1386$}} \\
Bengali  & \rrev{$+0.2616$} & \brev{\rrev{$+0.0997$}} \\
Hebrew   & \rrev{$+0.0724$} & \brev{\rrev{$0.0059$}} \\
Korean   & \rrev{$+0.0468$} & \brev{\rrev{$-0.0087$}} \\
\midrule
Mean (31 trained languages) & \rrev{$+0.0256$} & \brev{\rrev{$+0.0123$}} \\
\bottomrule
\end{tabular}
 \end{table}

\paragraph{Vocabulary entries that the language-model training data contains rarely.}
One candidate explanation for Tamil's rise is that the inverted-allocation tokenizer adds vocabulary entries for Tamil that occur too rarely in the language-model training data, which could leave their embeddings poorly trained.
\dealwithlater{The per-token losses of this analysis were recorded on the first 200 FLORES+ devtest sentences of one seed, so its Tamil rise differs slightly from the full-devtest values elsewhere in this appendix.} 
To test this, we \rrev{estimate}, how often each vocabulary entry occurs in a run's training-token stream of one inverted-allocation seed, then pair each per-token loss from that run's FLORES+ evaluation with its entry's training-frequency count.
Under the inverted-allocation tokenizer, $2.6\%$ of Tamil's FLORES+ token occurrences fall on entries that the training stream contains ten or fewer times vs. $0.0\%$ under the reference tokenizer.
These rare entries are indeed harder to predict: among Tamil entries of $15$ to $19$ bytes seen $100$ or fewer times cost $1.13$ bits per byte vs. $0.54$ for entries of the same length seen more than $1{,}000$ times.
But these entries do not account for Tamil's rise: only $3.9\%$ of the run's Tamil bits fall on them, and crediting them with every bit they receive accounts for at most $13.6\%$ of the run's Tamil rise of $+0.277$.
The rise is instead spread over the vocabulary: entries that are the same string under both tokenizers, with training-stream counts matched within a factor of $1.25$, cost $1.61$ times as many bits in Tamil under the inverted-allocation tokenizer as under the reference tokenizer.
\subsection{Exclusion and reallocation compared}
\label{app:sec:inversion:compared}
Excluding a language from the tokenizer-training data (\cref{sec:sensitivity}) and raising its share (\cref{sec:design}) change the same quantity in opposite directions, and both raise the language's FLORES+ BPB.
When a language is excluded, its text is segmented into more tokens, and its penalty grows with its fertility ratio.
When a language's share is raised, its text is segmented into fewer tokens, and the rise does not come from a smaller byte budget: under the inverted-allocation tokenizer, Tamil's share of the \rrev{seed-42} model's training bytes rose by $\rrev{0.015}$ percentage points while its FLORES+ BPB rose by $\rrev{+0.267}$\rrev{ (mean of two seeds)}.
Of Tamil's rise\rrev{ on repeated data}, matching its training-token count removes about a third (\cref{tab:tokenmatch-main}), and about half appears only with repeated data (\cref{tab:repetition}).

One mechanism that we hypothesize, which our experiments do not isolate, is overfitting of the token embeddings: the inverted-allocation tokenizers segment the low-resource languages into fewer, longer, more language-specific vocabulary entries (Tamil's fertility falls to $0.379$ times the reference tokenizer's; \cref{app:inversion}). These tokens necessarily occur in a smaller set of training contexts, and repeating those contexts may overfit the embeddings to them\clara{wording}.
A check \rrev{on per-vocabulary-entry losses (\cref{app:sec:inversion:further})} limits this explanation: the rise in loss is spread across vocabulary entries rather than concentrated on rarely trained ones.

\paragraph{\rrev{Tokenizer-training coverage without language-model training data raises the covered languages' BPB.}}
\rrev{The multilingual tokenizer-training data of four panel tokenizers is a sample of FineWeb2's full language pool; each of the four differs from a panel counterpart only in that data source (\cref{tab:grid}). Three of the four have a higher 31-language FLORES+ mean than their counterpart, by $+0.006$ to $+0.009$, and one a lower mean, by $0.008$; all four differences are beyond the two-run threshold. For the 183 evaluated languages outside the language-model training mixture, all four full-pool tokenizers have a higher mean FLORES+ BPB than their counterparts, by $+0.20$ to $+0.42$, and depending on the pair at most 56 of the 183 languages have a lower BPB under the full-pool tokenizer. These languages have vocabulary entries under the full-pool tokenizers, but the language-model training data contain no text for them, so the entries are rarely trained during language-model training. \Cref{sec:sensitivity} shows that removing a covered language from the tokenizer-training data raises its BPB; here, tokenizer-training coverage for a language absent from the language-model training data also raises its BPB. Coverage lowers a language's BPB only when the language model trains on that language.}

\section{Intrinsic Metrics: Definitions and Additional Results}
\label{app:extra-results}

\subsection{Intrinsic metric definitions}
\label{app:metrics}
Each metric is computed from a tokenizer applied to a text corpus, without training
a model; we measure each on the same FLORES+ text per language (and, in
\cref{app:fineweb2}, on FineWeb2 text). All values are computed with the
\tokeval{} library; each metric's exact definition is stated below.

\begin{itemize}[leftmargin=15pt,noitemsep]
\item \textbf{Compression rate}: UTF-8 bytes of text per token. Higher means each token covers more text.
\item \textbf{Fertility}: tokens per whitespace-delimited word. Higher means words
are split into more pieces. Fertility is closely related to compression rate (the
two are strongly correlated across tokenizers) and sits between an
information-theoretic measure of compression and a morphological measure of how
words are segmented.
\item \textbf{Unigram, bigram, and trigram entropy}: the Shannon entropy, in bits,
of the distribution of single tokens, token pairs, and token triples in the
tokenized corpus. Lower entropy means a more peaked, more predictable token
distribution.
\item \textbf{R\'enyi efficiency} ($\alpha = 2$): the order-2 R\'enyi entropy of the
unigram token distribution, normalized by $\log_2$ of the declared
vocabulary size, so that it lies in $[0,1]$.
Relative to Shannon entropy it down-weights the long tail of rare
tokens, and it was proposed as a predictor of language-model quality
\citep{zouhar-etal-2023-tokenization}. Values normalized instead by the number
of observed token types are on a different, per-language-incomparable scale and
cannot be compared with the values here.
\item \textbf{Vocabulary utilization}: the fraction of the vocabulary the tokenizer
actually uses on the corpus. Because it is the used-token count divided by
vocabulary size, it is not comparable across tokenizers of different vocabulary
sizes. One further caution for reuse: under the compression-rate covariate control
its within-language coefficient jumps from $+0.0002$ to
$+0.0105$ per standard deviation of the metric, a shift far larger than any other
metric's, so the controlled coefficient is a collinearity artifact with
compression rate and we do not interpret it.
\item \textbf{UTF-8 character split rate}: among the multi-byte UTF-8 characters in
the text, the fraction the tokenizer splits across a token boundary, so that one
character is spread over more than one token. It is well-defined only where
multi-byte characters are common.
\item \textbf{UTF-8 boundary-crossing rate}: the fraction of tokens whose byte span
crosses a UTF-8 character boundary (a token that begins or ends in the middle of a
character). Both UTF-8 rates are detected at the byte level as defined here. 
\end{itemize}

\begin{table}[t]
\centering
\caption{Within-language relationship between intrinsic metrics and FLORES BPB
on the 31 trained languages: $\beta$ from
$\text{BPB} \sim z(m) + (1\mid\text{language})$, in BPB per standard deviation of
the metric ($n=1674$ tokenizer--language pairs; $1673$ for UTF-8 char split, whose
value is unavailable for one (tokenizer, language) entry; 31 languages, 54
tokenizers; null-model language ICC $=\rrev{0.995}$).
SE$_{\text{m}}$: model-based standard error. SE$_{\text{cl}}$ and
$p_{\text{adj}}^{\text{cl}}$: cluster-robust standard error and BH-adjusted
$p$-value, clustered at the family level ($G = 42$ families); the clustered
$p$ is the inference we rely on. $R^2$: share of within-language variance
explained. Sorted by $|\beta|$. Vocabulary utilization's raw
$R^2$ is marginally negative, a degenerate output reported as $0.000$.
 All values are for models trained on the same number of tokens; \cref{tab:within-budget} reports the coefficients after the adjustment for the number of training bytes.}
\label{tab:within}
\small
\begin{tabular}{lccccc}
\toprule
\textbf{Intrinsic metric} & $\beta$ & SE$_{\text{m}}$ & SE$_{\text{cl}}$ & $p_{\text{adj}}^{\text{cl}}$ & $R^2$ \\
\midrule
Fertility               & \rrev{$+0.0156$} & \rrev{$0.0013$} & \rrev{$0.0027$} & \rrev{$<10^{-7}$} & \rrev{$0.082$} \\ 
UTF-8 char split        & \rrev{$+0.0113$} & \rrev{$0.0005$} & \rrev{$0.0012$} & $<10^{-16}$ & \rrev{$0.214$} \\ 
Trigram entropy         & \rrev{$-0.0112$} & \rrev{$0.0006$} & \rrev{$0.0015$} & $<10^{-13}$ & \rrev{$0.170$} \\ 
Bigram entropy          & \rrev{$-0.0083$} & $0.0006$ & \rrev{$0.0017$} & \rrev{$<10^{-5}$} & \rrev{$0.120$} \\ 
Compression rate        & \rrev{$-0.0072$} & \rrev{$0.0007$} & \rrev{$0.0024$} & $0.003$     & \rrev{$0.057$} \\ 
R\'enyi eff.\ ($\alpha{=}2$) & \rrev{$-0.0070$} & \rrev{$0.0005$} & \rrev{$0.0015$} & $<10^{-5}$ & \rrev{$0.093$} \\ 
Unigram entropy         & \rrev{$-0.0064$} & \rrev{$0.0006$} & $0.0021$ & \rrev{$0.003$} & \rrev{$0.056$} \\ 
UTF-8 boundary crossing & \rrev{$+0.0028$} & $0.0008$ & $0.0008$ & \rrev{$<10^{-3}$} & \rrev{$0.007$} \\ 
Vocab utilization       & \rrev{$+0.0013$} & $0.0010$ & \rrev{$0.0030$} & \rrev{$0.67$} & $0.000$ \\
\bottomrule
\end{tabular}
 \end{table}

\rrev{\Cref{tab:within} lists each metric's within-language coefficient on FLORES+ BPB, its model-based and family-clustered standard errors, and the adjusted $p$-value.}

\begin{figure}[t]
\centering
\includegraphics[width=0.8\textwidth]{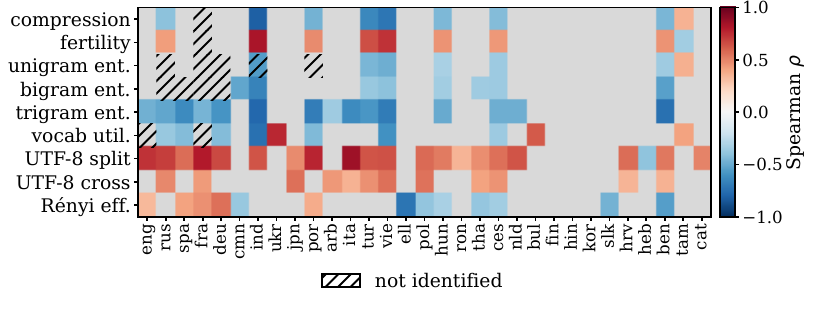}
\caption{Per-language Spearman $\rho$ between each intrinsic metric and FLORES+ BPB across the 54 tokenizers, for the 31 languages of the training mixture ordered from the highest to the lowest share of it.
Only BH-FDR-significant entries are colored.
The boundary-crossing entries for English, Indonesian, and Dutch are undefined (zero variance), and the char-split entry for Korean has $n = 53$.
Hatching marks the 13 entries that are not identified under the measured-composition control of \cref{app:extra-results}, where the metric and \rrev{the measured number of training bytes of the language} correlate at $|\rho| > 0.9$ across the tokenizer set.}
\label{fig:heatmap}
\end{figure}
\begin{figure}[h]
\centering
\includegraphics[width=0.8\textwidth]{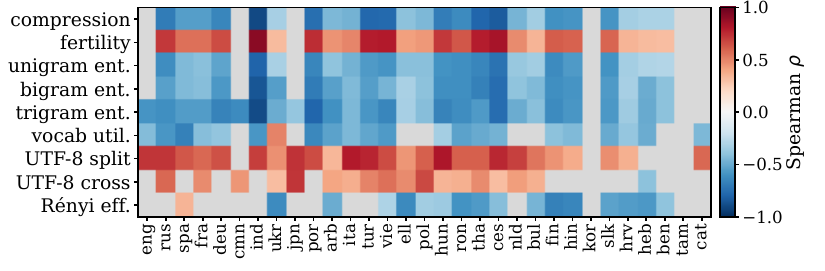}
\caption{Per-language Spearman $\rho$ between each FineWeb2-measured intrinsic
metric and per-language FineWeb2-validation BPB, across the 54 tokenizers, for
the 31 trained languages ordered from highest to lowest training-data share.
FineWeb2 analogue of \cref{fig:heatmap}.}
\label{fig:heatmap-fw2}
\end{figure}
\begin{figure}[h]
\centering
\includegraphics[width=0.8\textwidth]{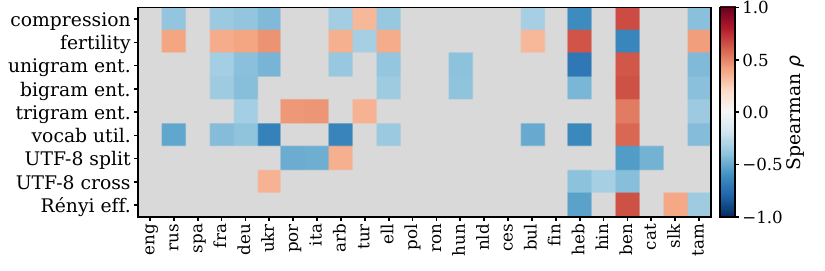}
\caption{Per-language Spearman $\rho$ between each intrinsic metric (measured on
FLORES+ text, as in \cref{fig:heatmap}) and per-language MultiBLiMP accuracy,
across the 54 tokenizers, for the 24 covered trained languages ordered from
highest to lowest training-data share.  The
boundary-crossing entries for English and Dutch are undefined (zero variance).}
\label{fig:heatmap-multiblimp}
\end{figure}

\paragraph{\rrev{Exemplars of \cref{fig:heatmap}.}}
As another example, three of the four metrics that correlate significantly with BPB for Tamil (compression rate, unigram entropy, and fertility) have correlation coefficients of the opposite sign from the mixed-effect model coefficients (\cref{tab:within}), which quantify the average within-language relationship between the metric and per-language BPB. This shows that single languages can benefit from tokenizer properties quite different from what helps on average.

\subsection{Per-language results on in-distribution FineWeb2 text and the full FLORES+ language set}
\label{app:fineweb2}

The main analyses correlate intrinsic metrics measured on FLORES+ with BPB
measured on FLORES+, which is parallel, curated text outside the training
distribution. 
This analysis looks at how results change when using in-distribution text for both measurements. Here, the intrinsic metrics are computed on FineWeb2 text, and the outcome is the
per-language FineWeb2-validation BPB, bits-per-byte on 100 held-out FineWeb2
documents. 
The section further looks at how results change when considering languages not explicitly in the model's training data mixture, i.e., when considering the full FLORES+ language set. 
These values are computed across all 54 tokenizers over the 31 trained languages.

\paragraph{The within-language relationships hold in-distribution and are generally stronger.}
\Cref{tab:within-fw2} repeats the within-language mixed effects model of \cref{tab:within} using FineWeb2 data. The coefficients of eight of the nine metrics have the same sign as in the FLORES+ analysis, and seven of these eight are larger in magnitude in-distribution: for example, fertility's coefficient is $+0.0322$ on FineWeb2 vs. $\rrev{+0.0156}$ on FLORES+.
Further, all nine metrics are significant when using the family-clustered standard error estimates on FineWeb2, vs. eight of nine on FLORES+.
Vocabulary utilization is the exception: it is not a significant predictor of FLORES+ BPB, but on FineWeb2 its coefficient is $-0.0076$ and significant, so tokenizers that use more of their vocabulary for a language have lower in-distribution BPB on that language.

\paragraph{Other per-language results also hold in-distribution.}
Of the 279 metric-by-language correlations, 211 are significant on FineWeb2 against \rrev{104} on FLORES+ (\cref{fig:heatmap-fw2}).
The correlation between a language's BPB spread across tokenizers and its log training-data weight remains negative and significant ($-0.394$ on FineWeb2 against $-0.541$ on FLORES+).
The mean off-diagonal Kendall $\tau$ between per-language tokenizer rankings is $0.455$ on FineWeb2 against $\rrev{0.418}$ on FLORES+ (\cref{sec:magnitude}), showing that disagreement between language-specific rankings is not specific to FLORES+; Ukrainian's agreement rises to $0.309$ and is no longer the lowest, suggesting that its extreme FLORES+ value is specific to that text.

\begin{table}[h]
\centering
\caption{Within-language relationship between intrinsic metrics and per-language
FineWeb2-validation BPB on the 31 trained languages, with both sides measured on
FineWeb2 text: $\beta$ from $\text{BPB} \sim z(m) + (1\mid\text{language})$, in
BPB per standard deviation of the metric ($n=1674$ tokenizer--language pairs; 31
languages, 54 tokenizers; null-model language ICC $=0.994$).
Columns as in \cref{tab:within}: model-based and family-clustered standard errors
($G=42$), BH-adjusted clustered $p$-value, share of within-language variance
explained. Sorted by $|\beta|$.}
\label{tab:within-fw2}
\small
\begin{tabular}{lccccc}
\toprule
\textbf{Intrinsic metric} & $\beta$ & SE$_{\text{m}}$ & SE$_{\text{cl}}$ & $p_{\text{adj}}^{\text{cl}}$ & $R^2$ \\
\midrule
Fertility               & $+0.0322$ & $0.0014$ & $0.0022$ & $<10^{-47}$ & $0.244$ \\
Trigram entropy         & $-0.0168$ & $0.0005$ & $0.0016$ & $<10^{-26}$ & $0.458$ \\
Unigram entropy         & $-0.0151$ & $0.0005$ & $0.0020$ & $<10^{-13}$ & $0.329$ \\
Compression rate        & $-0.0151$ & $0.0007$ & $0.0027$ & $<10^{-7}$  & $0.240$ \\
Bigram entropy          & $-0.0145$ & $0.0004$ & $0.0017$ & $<10^{-17}$ & $0.396$ \\
UTF-8 char split        & $+0.0140$ & $0.0005$ & $0.0008$ & $<10^{-71}$ & $0.344$ \\
R\'enyi eff.\ ($\alpha{=}2$) & $-0.0121$ & $0.0005$ & $0.0015$ & $<10^{-14}$ & $0.298$ \\
Vocab utilization       & $-0.0076$ & $0.0010$ & $0.0008$ & $<10^{-18}$ & $0.035$ \\
UTF-8 boundary crossing & $+0.0022$ & $0.0007$ & $0.0010$ & $0.026$     & $0.005$ \\
\bottomrule
\end{tabular}
 \end{table}

\paragraph{Results on entire FLORES+ dataset.}\label{app:full-flores}

At the aggregate level, for most metrics the sign of the correlation between the
tokenizer's global metric value and its mean per-language FLORES+ BPB flips when
the mean is taken over all 214 FLORES+ languages instead of the 31 trained
languages (\cref{tab:full-flores}); R\'enyi efficiency is the exception. 
There are several possible reasons that could explain this effect. 
The 214-language mean is a mean over 31 trained and 183 untrained languages, so the untrained languages contribute most of its terms. 
\clara{READ + POLISH}As the median columns of \cref{tab:full-flores} show, the per-language correlations between a tokenizer's global metric value and one language's FLORES+ BPB have, for most metrics, one sign on the trained languages and the opposite sign on the untrained languages; the aggregate correlation follows whichever set the mean is taken over, and no single language's relationship changes. R\'enyi efficiency is the exception: its aggregate correlation keeps its sign on both language sets.
The direction of the change is consistent with the finite-vocabulary trade-off that motivates this study: the tokenizers with the best global metric values on the trained mixture are those whose vocabularies fit the trained languages most closely, and \cref{sec:sensitivity} shows that a language absent from the tokenizer-training data has higher BPB. The 183 untrained languages are absent from both the tokenizer-training and the language-model training data.
Two cautions apply. The untrained-language correlations use the tokenizer's global metric value, not a measurement on the untrained language's own text. And the per-language seed variation over the 214 languages is $0.0146$, about six times the $0.0023$ of the trained languages (\cref{app:methods}), so the individual untrained-language correlations are noisier than their trained-language counterparts. 

\begin{table}[h]
\centering
\caption{Per-language decomposition of the full FLORES+ language set, across
the 54 tokenizers. Agg.\ $\rho$: Spearman between the tokenizer's global metric
value and its mean per-language FLORES+ BPB over the 31 trained languages
(tr.)\ and over all 214 languages. Median lang.\ $\rho$: median of the
per-language correlations (global metric value against one language's BPB)
over the trained and the untrained languages. Sig.: languages with a significant
per-language correlation (BH-FDR within metric), of 31 trained and of 183
untrained.}
\label{tab:full-flores}
\small
\begin{tabular}{lcccccc}
\toprule
& \multicolumn{2}{c}{\textbf{Agg.\ $\rho$}} & \multicolumn{2}{c}{\textbf{Median lang.\ $\rho$}} & \multicolumn{2}{c}{\textbf{Sig.}} \\
\textbf{Intrinsic metric} & \textbf{tr.-31} & \textbf{all-214} & \textbf{trained} & \textbf{untrained} & \textbf{tr.} & \textbf{untr.} \\
\midrule
R\'enyi eff.\ ($\alpha{=}2$) & \rrev{$-0.455$} & \rrev{$-0.193$} & \rrev{$-0.288$} & \rrev{$-0.072$} & \rrev{9} & \rrev{11} \\ 
Trigram entropy              & \rrev{$-0.446$} & \rrev{$+0.130$} & \rrev{$-0.328$} & \rrev{$+0.080$} & \rrev{15} & \rrev{34} \\ 
Compression rate             & \rrev{$-0.326$} & \rrev{$+0.227$} & \rrev{$-0.226$} & \rrev{$+0.106$} & \rrev{9} & \rrev{56} \\ 
UTF-8 char split             & \rrev{$+0.331$} & \rrev{$-0.194$} & \rrev{$+0.348$} & \rrev{$-0.047$} & \rrev{17} & \rrev{34} \\ 
Bigram entropy               & \rrev{$-0.286$} & $+0.052$ & \rrev{$-0.214$} & \rrev{$+0.024$} & \rrev{6} & \rrev{40} \\ 
Unigram entropy              & \rrev{$-0.274$} & \rrev{$+0.100$} & \rrev{$-0.224$} & \rrev{$+0.068$} & \rrev{5} & \rrev{50} \\ 
UTF-8 boundary crossing      & \rrev{$+0.266$} & \rrev{$-0.126$} & \rrev{$+0.273$} & \rrev{$0.005$} & \rrev{5} & \rrev{10} \\ 
Vocab utilization            & \rrev{$-0.040$} & \rrev{$+0.245$} & \rrev{$-0.104$} & \rrev{$+0.079$} & \rrev{2} & \rrev{33} \\ 
Fertility                    & \rrev{$+0.033$} & \rrev{$-0.303$} & \rrev{$+0.008$} & \rrev{$-0.125$} &  1 & \rrev{28} \\ 
\bottomrule
\end{tabular}
 \end{table}

\subsection{Controlling for confounders}
\label{app:methods:bytes}
\begin{table}[h]
\centering
\caption{\brev{Within-language coefficients of \cref{tab:within} after each run's BPB is adjusted for its number of training bytes (\rrev{\cref{app:methods:bytes}}), for the 51 tokenizers with a measured number of training bytes.
In each column the adjustment is made with one estimate of $\beta$, the change in the 31-language FLORES+ mean per unit increase in the natural logarithm of the number of training bytes: from left to right, no adjustment, the SuperBPE byte-matched retrain, the thirteen repeated-data continuations, the UnigramLM byte-matched retrain, and the eleven single-pass continuations.
$^{*}$: BH-adjusted family-clustered $p < 0.05$.
The unadjusted column is the fit of \cref{tab:within} on the same 51 tokenizers, so its values differ slightly from that table, which uses all 54.}}
\label{tab:within-budget}
\small
\begin{tabular}{lrrrrr}
\toprule
& \multicolumn{1}{c}{\textbf{unadjusted}} & \multicolumn{4}{c}{\textbf{estimate of $\beta$ used in the adjustment}} \\
\cmidrule(lr){3-6}
\textbf{Metric} & & $-0.0186$ & $-0.0257$ & \brev{$-0.0365$} & $-0.0397$ \\
\midrule
Fertility & \rrev{$+0.0154^{*}$} & \brev{\rrev{$+0.0154^{*}$}} & \brev{\rrev{$+0.0147^{*}$}} & \brev{\rrev{$+0.0135^{*}$}} & \brev{\rrev{$+0.0135^{*}$}} \\
UTF-8 char split & \rrev{$+0.0113^{*}$} & \rrev{$+0.0113^{*}$} & \brev{$+0.0108^{*}$} & \rrev{$+0.0099^{*}$} & \rrev{$+0.0099^{*}$} \\
Trigram entropy & \rrev{$-0.0110^{*}$} & \brev{\rrev{$-0.0109^{*}$}} & \brev{\rrev{$-0.0095^{*}$}} & \brev{\rrev{$-0.0070^{*}$}} & \brev{\rrev{$-0.0072^{*}$}} \\
Bigram entropy & $-0.0083^{*}$ & \brev{\rrev{$-0.0082^{*}$}} & \brev{\rrev{$-0.0067^{*}$}} & \brev{\rrev{$-0.0039^{*}$}} & \brev{$-0.0041^{*}$} \\
Compression rate & \rrev{$-0.0071^{*}$} & \brev{\rrev{$-0.0070^{*}$}} & \brev{\rrev{$-0.0055$}} & \brev{\rrev{$-0.0029$}} & \brev{\rrev{$-0.0031$}} \\
R\'enyi eff.\ ($\alpha{=}2$) & \rrev{$-0.0071^{*}$} & \brev{\rrev{$-0.0070^{*}$}} & \brev{\rrev{$-0.0055^{*}$}} & \brev{\rrev{$-0.0027^{*}$}} & \brev{\rrev{$-0.0029$}} \\
Unigram entropy & \rrev{$-0.0065^{*}$} & \brev{\rrev{$-0.0064^{*}$}} & \brev{\rrev{$-0.0050^{*}$}} & \brev{\rrev{$-0.0026$}} & \brev{\rrev{$-0.0028$}} \\
UTF-8 boundary crossing & \rrev{$+0.0025^{*}$} & \rrev{$+0.0025^{*}$} & \rrev{$+0.0023^{*}$} & \brev{\rrev{$+0.0020^{*}$}} & \rrev{$+0.0020^{*}$} \\
Vocab.\ utilization & \rrev{$+0.0011$} & \brev{\rrev{$+0.0012$}} & \brev{\rrev{$+0.0022$}} & \brev{\rrev{$+0.0041$}} & \brev{\rrev{$+0.0040$}} \\
\bottomrule
\end{tabular}
 \end{table}

\begin{table}[h]
\centering
\caption{Within-language coefficient $\beta$ of each intrinsic metric on FLORES+ BPB, fit separately within resource tiers (high $n{=}6$, mid $n{=}15$, low $n{=}10$ languages, by share of the language-model training data), in BPB per standard deviation of the metric.
Sorted by the low-tier $|\beta|$. The coefficients are the mixed-effects tier fits.
\textbf{Bold}: BH-adjusted $p_{\text{adj}} < 0.05$ under the family-clustered tier fit, an OLS fit with language fixed effects clustered by family ($G = 42$; \cref{app:methods}); \rrev{the compression-rate high-tier entry is not significant ($p_{\text{adj}} = 0.063$)}\brev{, and it is also not significant after the adjustment for the number of training bytes (\rrev{\cref{app:methods:bytes}})}.
}
\label{tab:tier}
\small
\begin{tabular}{lccc}
\toprule
\textbf{Intrinsic metric} & \textbf{High} & \textbf{Mid} & \textbf{Low} \\
\midrule
Trigram entropy              & \rrev{$\mathbf{-0.0111}$} & \rrev{$\mathbf{-0.0084}$} & \rrev{$\mathbf{-0.0128}$} \\ 
UTF-8 char split             & \rrev{$\mathbf{+0.0215}$} & \rrev{$\mathbf{+0.0074}$} & \rrev{$\mathbf{+0.0119}$} \\ 
Vocab utilization            & \rrev{$-0.0052$} & \rrev{$+0.0012$} & \rrev{$\mathbf{+0.0112}$} \\ 
R\'enyi eff.\ ($\alpha{=}2$) & \rrev{$-0.0002$} & \rrev{$\mathbf{-0.0065}$} & \rrev{$\mathbf{-0.0105}$} \\ 
Fertility                    & \rrev{$\mathbf{+0.0539}$} & \rrev{$\mathbf{+0.0070}$} & \rrev{$\mathbf{+0.0099}$} \\ 
Bigram entropy               & \rrev{$-0.0060$} & \rrev{$\mathbf{-0.0077}$} & \rrev{$\mathbf{-0.0094}$} \\ 
Unigram entropy              & \rrev{$-0.0062$} & \rrev{$\mathbf{-0.0059}$} & \rrev{$-0.0059$} \\ 
Compression rate             & \rrev{$-0.0103$} & \rrev{$\mathbf{-0.0063}$} & \rrev{$-0.0043$} \\ 
UTF-8 boundary crossing      & \rrev{$\mathbf{+0.0101}$} & \rrev{$\mathbf{+0.0028}$} & \rrev{$+0.0028$} \\ 
\bottomrule
\end{tabular}
 \end{table}

 The training budget is fixed in tokens, so a model whose tokenizer compresses a language better is trained on more bytes of that language. 
For a fixed token budget, several metrics (e.g., compression rate and fertility) are therefore directly related to how much training bytes a model will see.  
This could be a confounder in any relationship we observe between one of these metrics and BPB. 
We measure each run's number of training bytes of each language with a census of the training data for 51 of the 54 tokenizers.
The three tokenizers that use the encoding based on Unicode script classes (\cref{sec:setup}) train on pre-tokenized shards, for which we did not record the number of bytes; they are excluded from both analyses below.
We use the measurement as an offset to the pooled within-language coefficients of \cref{tab:within}, and as a covariate for the per-language correlations of \cref{fig:heatmap}.

\paragraph{Pooled coefficients adjusted for training bytes.}
Let $x$ be the natural logarithm of a run's number of training bytes of the language, $\bar{x}$ its mean, and $\beta$ an estimate of the change in BPB per unit increase in $x$. We replace each per-language BPB of a run by $\text{BPB} - \beta\,(x - \bar{x})$, with the same $\beta$ for every run and every language, and refit the within-language models of \cref{sec:setup} and the tier fits of \cref{sec:metrics}.
We do not fit $\beta$ in these models; instead we take it from runs that differ from their counterparts only in the number of training bytes. Several of these runs exist from various ablations performed throughout the study: the byte-matched retrains of the SuperBPE and UnigramLM tokenizers, the thirteen repeated-data continuations, and the eleven single-pass continuations.
This gives four estimates of $\beta$, the largest about twice the smallest, and \cref{tab:within-budget} reports the refit under each. \dealwithlater{The $p$-values of that table do not include the uncertainty of $\beta$.}
The adjustment does not change the sign of any coefficient. Fertility, the UTF-8 character-split rate, bigram and trigram entropy, and the UTF-8 boundary-crossing rate remain significant under all four estimates. Compression rate is significant under only the smallest estimate, and unigram entropy and Rényi efficiency lose significance under the larger estimates.
The relationships that remain are those of metrics that describe how text is segmented; compression rate, the metric most directly tied to the number of training bytes, does not retain a within-language relationship with BPB once that number is adjusted for. The tier fits of \cref{tab:tier} change in the same way.

\paragraph{Per-language correlations given the language's training bytes.} For each entry of \cref{fig:heatmap}, we recompute the Spearman correlation between the metric and the language's BPB as a partial correlation given the run's number of training bytes of the language, over the 51 tokenizers.
An entry is not identified when the metric and the number of training bytes correlate at $|\rho| > 0.9$ across the tokenizers, because the partial correlation cannot then separate them.
Thirteen of the 276 entries are not identified, and they are hatched in \cref{fig:heatmap}.
Of the identified entries that are significant at tokenizer level, about three-fifths remain significant under the token count partial correlation, and about one in ten change sign. 
\rrev{The sign changes are concentrated in specific languages: four are Thai entries, three Hungarian, two Chinese, one Arabic, and one Czech, and nine of the eleven are entries of the entropy metrics or of R\'enyi efficiency. The entries that lose significance also concentrate by metric: one of unigram entropy's six significant entries remains significant under the partial correlation, two of bigram entropy's eight, sixteen of the character-split rate's twenty-one, and twelve of trigram entropy's fifteen.}
The reasons behind the sensitivity of these measurements is an investigation we leave for future work. 

\paragraph{R\'enyi efficiency is not significantly related to FLORES+ BPB for the languages with the most language-model training data.}
We ask whether a metric's within-language coefficient depends on the language's share of the language-model training data.
We order the 31 languages by this share, form groups of high-resource, mid-resource, and low-resource languages, and refit the within-language model within each group (\cref{tab:tier} in \cref{app:extra-results}).
The coefficient of R\'enyi efficiency is $-0.0009$ and not significant in the high-resource group, and $-0.0116$ and significant in the low-resource group.\footnote{We standardize each metric within its group before the fit, so the magnitudes of the coefficients are not directly comparable between groups.}
We also fit one model per metric on all 31 languages, with an interaction between the metric and the logarithm of the language's share.
For R\'enyi efficiency the interaction is significant ($+0.0061$), so its negative coefficient is closer to zero for languages with a larger share. \rrev{Higher R\'enyi efficiency goes together with lower BPB mainly for the languages with small shares of the language-model training data; for the highest-share languages the association is near zero.} 
For the character-split rate the interaction is also significant ($+0.0165$), so its positive coefficient is larger for languages with a larger share.
For the other seven metrics, fertility among them, the interaction is not significant.

\dealwithlater{\paragraph{Vocabulary utilization has no pooled within-language relationship with BPB, and its per-language relationship changes sign.}
Its per-language correlation spans $-0.75$ (Indonesian) to $+0.74$ (Ukrainian), both significant,
and its tier coefficients have opposite signs: $-0.0066$ at high resource and $+0.0116$ at low resource.
Under the family-clustered inference of the tier fits, the low-tier coefficient is significant and the high-tier coefficient is not (\cref{tab:tier}).
For reuse
beyond these 54 tokenizers, one caveat: because utilization is the used-token count divided by
vocabulary size, it depends on vocabulary size and sample size, so it is not
comparable across tokenizers of different vocabulary sizes.}

\subsection{Tokenizer ranking}
\label{app:proxy-details}
\begin{table}[t]
\centering
\caption{Pairwise accuracy at ordering held-out tokenizer pairs by the 1.27B models' aggregate BPB, under leave-one-family-out cross-validation over the 42 tokenizer families ($40$ for the validation-BPB target, which excludes the three SCRIPT tokenizers; \cref{app:methods}).
Val BPB: validation BPB. FLORES-tr.: the 31-language FLORES+ mean.
Tokenizer-level rows (second block) use one value of each metric per tokenizer. Per-language rows (third block) use the same metrics measured on each of the 31 languages ($278$ features; \cref{app:proxy-details} states the excluded column);  a single-metric row of that block uses all per-language measurements of that metric as the features.
The majority-class baseline is the accuracy of always predicting the more frequent pair order.
Each single-metric row uses one metric fixed in advance; selecting the best single-metric row from this table afterward is a choice made in hindsight, so we recommend the fixed nine-metric combination for screening.
The three smaller-model rows are direct ranking transfer, not cross-validation: the 17 tokenizers with runs at all four scales are ranked by the smaller model's validation BPB and the fraction of pairs ordered as by the 1.27B outcome is reported\dealwithlater{ (\cref{app:training} discloses a d16 weight-decay inconsistency)}.}
\label{tab:proxy}
\small
\begin{tabular}{lcc}
\toprule
\textbf{Ranking method} & \textbf{Val BPB} & \textbf{FLORES-tr.} \\
\midrule
Majority-class baseline                & \brev{$0.601$} & $0.565$ \\
\midrule
All nine metrics (logistic ridge)      & \brev{$0.700$} & \rrev{$0.735$} \\
Best single metric, selected per fold  & \brev{$0.655$} & $0.668$ \\
Bigram entropy (fixed)                 & \brev{$0.683$} & \rrev{$0.676$} \\
Trigram entropy (fixed)                & \brev{$0.639$} & $0.664$ \\
R\'enyi eff.\ ($\alpha{=}2$, fixed)    & \brev{$0.675$} & $0.653$ \\
Compression rate (fixed)               & \brev{$0.669$} & $0.657$ \\
\midrule
All nine metrics, per language ($278$ features) & \brev{$0.907$} & \rrev{$0.890$} \\
Best single feature, selected per fold        & \brev{$0.848$} & \rrev{$0.799$} \\
Bigram entropy block (fixed)                & \brev{$0.910$} & \rrev{$0.909$} \\
Trigram entropy block (fixed)               & \brev{$0.879$} & \rrev{$0.865$} \\
R\'enyi eff.\ block ($\alpha{=}2$, fixed)   & \brev{$0.860$} & \rrev{$0.824$} \\
Compression rate block (fixed)              & \brev{$0.828$} & \rrev{$0.832$} \\
\midrule
Smaller model, d8 (val-BPB rank, $n{=}17$)  & \brev{$0.868$} & \rrev{$0.875$} \\
Smaller model, d12 (val-BPB rank, $n{=}17$) & \brev{$0.846$} & \rrev{$0.853$} \\
Smaller model, d16 (val-BPB rank, $n{=}17$) & \brev{$0.868$} & \rrev{$0.831$} \\
\bottomrule
\end{tabular}
 \end{table}
\paragraph{Implementation.}
To predict which of two tokenizers will have better downstream BPB, we train logistic pairwise rankers. 
For each pair of tokenizers $(a,b)$ in a fixed canonical order, the feature vector is the difference of the two tokenizers' metric vectors, and the label is whether $b$'s aggregate BPB exceeds $a$'s.
Features are z-scored (standardized to mean zero and standard deviation one) using statistics computed on the training pairs only.
The classifier is scikit-learn \texttt{LogisticRegression} with an L2 penalty, the \texttt{lbfgs} solver, and an intercept; the prediction for a held-out pair is the sign of the fitted linear score. 
The ridge strength $C$ is selected inside each training fold by an inner tokenizer family-wise cross-validation over a 13-point logarithmic grid from $10^{-3}$ to $10^{3}$.
The procedure is deterministic and uses no random seeds.
The tokenizer-level feature space has one value of each of the nine metrics per tokenizer.
The per-language feature space has the same nine metrics measured on each of the 31 languages: $9 \times 31 = 279$ columns, minus the Korean character-split column, which is unavailable for one tokenizer (\cref{fig:heatmap}), leaving $278$.
To account for dependence among closely related tokenizers, we use leave-one-family-out cross-validation.
In each fold, we hold out an entire tokenizer family, so that a tokenizer is never ranked by a model trained on another member of its own family. 
The 31-language FLORES+ mean target uses all 54 tokenizers (42 families); the validation-BPB target excludes the three SCRIPT tokenizers, for which no validation BPB exists (\cref{app:methods}), leaving 51 tokenizers in 40 families.
We evaluate pairwise accuracy, the fraction of evaluated tokenizer pairs whose predicted ordering matches their ordering according to the trained models' BPB.
We score every pair for which at least one tokenizer belongs to the held-out
family.

\paragraph{Per-language ranking accuracies.}
For each of the 31 languages, we rank the 54 tokenizers by that language's FLORES+ BPB and score the ridge combination of the nine metrics on held-out pairs under the same leave-one-family-out folds as \cref{tab:proxy}.
In the per-language feature space, the nine columns measured on the target language itself are dropped before fitting, so a language's ranking is never predicted from metrics measured on its own text.
With the metrics as one value per tokenizer, the per-language accuracies have mean $0.713$ and median $0.709$ and range from $0.615$ (Croatian) to $0.802$ (Tamil) vs. $0.735$ for the 31-language FLORES+ mean target.
With the metrics measured per language, the accuracies have mean $0.829$ and median $0.837$ and range from $0.738$ (Czech) to $0.878$ (Tamil) vs. $\rrev{0.890}$ for the same target.

\section{Task accuracies as the outcome: MultiBLiMP, XNLI, XCOPA, TokSuite, and XWinograd}
\label{app:multiblimp}

We repeat the within-language analyses with per-language accuracy on MultiBLiMP, XNLI, XCOPA, TokSuite, and XWinograd in place of BPB.
\rrev{XCOPA is a two-alternative causal-commonsense benchmark covering 7 of the 31 languages; XWinograd is a two-alternative coreference benchmark covering 6; the canonical subset of TokSuite is a four-way multiple-choice benchmark covering 4.}

\paragraph{MultiBLiMP.}
MultiBLiMP accuracy (\cref{sec:setup}) is the fraction of minimal pairs whose grammatical member the model assigns higher probability. It is available for 24 of the 31 trained languages and all 54 tokenizers.
Per-language accuracies range from $0.667$ to $1.0$; both endpoints are Bengali, which has only 21 minimal pairs and therefore the noisiest estimates in our language set.
For 16 of the 24 languages, the pooled seed standard deviation does not exceed the binomial standard error implied by the benchmark item count, so the uncertainty from the finite benchmark is at least as large as the observed run-to-run variation; these are the \brev{open dots} in \cref{fig:magnitude}.

\Cref{tab:multiblimp} repeats the within-language model of \cref{sec:setup} with MultiBLiMP accuracy as the outcome. Eight of the nine intrinsic metrics are significant under the family-clustered inference, with vocabulary utilization again the exception.
At the individual-language level, 67 of the 216 metric-by-language correlations are significant after adjustment across all entries (\cref{fig:heatmap-multiblimp}).

\begin{table}[h]
\centering
\caption{Within-language relationship between intrinsic metrics and per-language
MultiBLiMP accuracy on the 24 covered trained languages: $\beta$ from
$\text{accuracy} \sim z(m) + (1\mid\text{language})$, in accuracy per standard
deviation of the metric ($n=1296$ tokenizer--language pairs; 24 languages, 54
tokenizers).
SE$_{\text{m}}$, SE$_{\text{cl}}$, and $p_{\text{adj}}^{\text{cl}}$ as in
\cref{tab:within} ($G = 42$ families; the clustered $p$-value is the inference we
rely on). Sorted by $|\beta|$.}
\label{tab:multiblimp}
\small
\begin{tabular}{lcccc}
\toprule
\textbf{Intrinsic metric} & $\beta$ & SE$_{\text{m}}$ & SE$_{\text{cl}}$ & $p_{\text{adj}}^{\text{cl}}$ \\
\midrule
UTF-8 char split        & $-0.0066$ & $0.0006$ & $0.0014$ & $<10^{-5}$ \\
Fertility               & $-0.0056$ & $0.0007$ & $0.0012$ & $<10^{-5}$ \\
UTF-8 boundary crossing & $-0.0046$ & $0.0009$ & $0.0009$ & $<10^{-5}$ \\
Trigram entropy         & $+0.0040$ & $0.0007$ & $0.0010$ & $<10^{-4}$ \\
Bigram entropy          & $+0.0035$ & $0.0006$ & $0.0009$ & $<10^{-3}$ \\
R\'enyi eff.\ ($\alpha{=}2$) & $+0.0031$ & $0.0006$ & $0.0007$ & $<10^{-4}$ \\
Unigram entropy         & $+0.0029$ & $0.0007$ & $0.0009$ & $0.003$ \\
Compression rate        & $+0.0029$ & $0.0008$ & $0.0009$ & $0.004$ \\
Vocab utilization       & $-0.0002$ & $0.0011$ & $0.0010$ & $0.64$ \\
\bottomrule
\end{tabular}
 \end{table}

\paragraph{XNLI.}
XNLI~\citep{conneau2018xnli} is a three-way natural-language-inference benchmark (entailment, neutral, contradiction) scored by loglikelihood, with chance accuracy $1/3$. We use the 13 of its 15 languages that are among the 31 trained languages (dropping Swahili and Urdu), with $2{,}490$ items per language, and evaluate all 54 tokenizers.
Across the 54 tokenizers, per-language accuracies span $0.323$ to $0.512$, and several languages stay close to chance for every tokenizer (Arabic peaks at $0.355$, Chinese at $0.374$).

Mean XNLI accuracy correlates with the 31-language FLORES+ mean at Spearman $\rho=-0.31$ ($n=54$, significant) and with validation BPB at $\rho=-0.26$ ($n=54$, not significant).
Some per-language intrinsic-metric relationships survive on XNLI, though fewer than on MultiBLiMP: 16 of the 117 metric-by-language correlations are significant after BH adjustment, and four metrics remain significant in the family-clustered within-language model ($G=42$ families): fertility, trigram entropy, UTF-8 boundary crossing and UTF-8 character splitting (\cref{tab:xnli}).

\begin{table}[h]
\centering
\caption{Within-language relationship between intrinsic metrics and per-language
XNLI accuracy on the 13 covered trained languages: $\beta$ from
$\text{accuracy} \sim z(m) + (1\mid\text{language})$, in accuracy per standard
deviation of the metric ($n=702$ tokenizer--language pairs; 13 languages, 54
tokenizers).
SE$_{\text{m}}$, SE$_{\text{cl}}$, and $p_{\text{adj}}^{\text{cl}}$ as in
\cref{tab:within} ($G = 42$ families; the clustered $p$-value is the inference we
rely on). Sorted by $|\beta|$.}
\label{tab:xnli}
\small
\begin{tabular}{lcccc}
\toprule
\textbf{Intrinsic metric} & $\beta$ & SE$_{\text{m}}$ & SE$_{\text{cl}}$ & $p_{\text{adj}}^{\text{cl}}$ \\
\midrule
Fertility               & $-0.0043$ & $0.0012$ & $0.0009$ & $<10^{-5}$ \\
Trigram entropy         & $+0.0042$ & $0.0008$ & $0.0010$ & $<10^{-4}$ \\
UTF-8 char split        & $-0.0030$ & $0.0006$ & $0.0007$ & $<10^{-4}$ \\
Compression rate        & $+0.0023$ & $0.0008$ & $0.0011$ & $0.07$ \\
UTF-8 boundary crossing & $-0.0020$ & $0.0006$ & $0.0004$ & $<10^{-5}$ \\
Unigram entropy         & $+0.0019$ & $0.0007$ & $0.0010$ & $0.08$ \\
Bigram entropy          & $+0.0019$ & $0.0006$ & $0.0010$ & $0.07$ \\
Vocab utilization       & $+0.0010$ & $0.0010$ & $0.0011$ & $0.52$ \\
R\'enyi eff.\ ($\alpha{=}2$) & $+0.0006$ & $0.0006$ & $0.0011$ & $0.58$ \\
\bottomrule
\end{tabular} \end{table}

\paragraph{XCOPA.}
XCOPA \citep{ponti-etal-2020-xcopa} is a two-alternative causal-commonsense benchmark scored by loglikelihood, with chance accuracy $0.5$. It covers seven of the 31 trained languages, with $500$ items per language, and we evaluate all 54 tokenizers and the five allocation runs of \cref{app:inversion}.
Across the 54 tokenizers, per-language accuracies span only $0.468$ to $0.586$.
Nevertheless, mean XCOPA accuracy is associated with the BPB outcomes: it correlates with validation BPB at Spearman $\rho=-0.49$ ($n=51$) and with the 31-language FLORES+ mean at $\rho=-0.52$ ($n=54$); both correlations are significant.

The per-language intrinsic-metric relationships are much weaker on XCOPA: only 3 of the 63 metric-by-language correlations are significant after BH adjustment, and no metric is significant in the family-clustered within-language model ($G=42$ families).
Five of the seven languages have a cross-tokenizer accuracy spread larger than their seed variation. For all seven, however, the pooled seed standard deviation is no larger than the binomial standard error at $500$ items, so benchmark sampling uncertainty is at least as large as the observed run-to-run variation.
\paragraph{TokSuite.}
TokSuite~\citep{altintas2026toksuite} is a four-way multiple-choice benchmark on general-knowledge questions and their surface-level variations, scored by loglikelihood, with chance accuracy $0.25$. We use the four languages that are also among our trained languages (English, Turkish, Italian and Chinese), pooling the canonical and perturbed subsets for $485$ to $1{,}140$ items per language, and we evaluate all 54 tokenizers.
Across the 54 tokenizers, per-language accuracies span $0.206$ to $0.678$. English is the only language well above chance on average ($0.494$, against $0.30$--$0.32$ for the other three).
Mean TokSuite accuracy shows no association with either BPB outcome: $\rho=-0.02$ with validation BPB and $\rho=-0.05$ with the 31-language FLORES+ mean ($n=54$, neither significant).
The per-language intrinsic-metric relationships are similarly weak: only 2 of the 36 metric-by-language correlations are significant after BH adjustment, and no metric is significant in the family-clustered within-language model ($G=42$ families).

\paragraph{XWinograd.}
XWinograd~\citep{tikhonov-ryabinin-2021-heads,muennighoff-etal-2023-crosslingual} is a two-alternative Winograd-schema coreference benchmark scored by loglikelihood, with chance accuracy $0.5$. All six of its languages (English, French, Japanese, Portuguese, Russian and Chinese) are among the 31 trained languages, with between $83$ (French) and $2{,}325$ (English) items per language, and we evaluate all 54 tokenizers.
Across the 54 tokenizers, per-language accuracies span $0.434$ to $0.680$; only English, Russian and Chinese average clearly above chance, and Japanese stays within $0.04$ of chance for every tokenizer.
Mean XWinograd accuracy is not significantly associated with either BPB outcome ($\rho=-0.12$ with validation BPB, $\rho=-0.16$ with the 31-language FLORES+ mean; $n=54$).

\paragraph{Belebele.}
We also evaluated 27 of the 54 runs on Belebele~\citep{bandarkar-etal-2024-belebele}.
Across these runs, aggregate accuracy ranges from $0.2259$ to $0.2715$, compared with a chance level of $0.25$.
Because performance remains approximately at chance, we do not use Belebele in our analyses.

\end{document}